\documentclass[pdflatex,sn-basic]{sn-jnl}% Basic Springer Nature Reference Style/Chemistry Reference Style

\jyear{2024}

\usepackage{textcomp}
\usepackage{hyperref}
\theoremstyle{thmstyleone}

\theoremstyle{thmstyletwo}

\theoremstyle{thmstylethree}

\usepackage{lineno}
\usepackage{soul}
\usepackage{graphicx}
\usepackage{array,tabularx,booktabs,longtable}
\graphicspath{ {./} }
\usepackage{makecell}
\usepackage{pifont}
\usepackage{url}
\usepackage{adjustbox}
\usepackage{pdflscape}
\usepackage{longtable}

\usepackage{anyfontsize}

\definecolor{blue}{rgb}{0,0,1}

\newcommand{\rd}[1]{{\color{black}#1}}
\newcommand{\xmark}{\ding{55}}%

\begin{document}

\title[Ultrasound-Based AI for COVID-19 Detection]{Ultrasound-Based AI for COVID-19 Detection: A Comprehensive Review of Public and Private Lung Ultrasound Datasets and Studies}

\author[1]{\fnm{Abrar} \sur{Morshed}}\email{abrar03.morshed@gmail.com}
\equalcont{These authors contributed equally to this work.}
\author[2]{\fnm{Abdulla} \sur{Al Shihab}}\email{Abdullahshihab54321@gmail.com}
\equalcont{These authors contributed equally to this work.}
\author*[3,4]{\fnm{Md Abrar} \sur{Jahin}}\email{abrar.jahin.2652@gmail.com}
\cofirst{Co-first Authors.}
\author[5]{\fnm{Md Jaber} \sur{Al Nahian}}\email{nahianrism@gmail.com}
\equalcont{These authors contributed equally to this work.}
\author[6]{\fnm{Md Murad Hossain} \sur{Sarker}}\email{mh6367828@gmail.com}
\equalcont{These authors contributed equally to this work.}
\author[7]{\fnm{Md Sharjis} \sur{Ibne Wadud}}\email{sharjis@du.ac.bd}
\cofirst{Co-first Authors.}
\author[8]{\fnm{Mohammad Istiaq} \sur{Uddin}}\email{mistiaq@ieee.org}
\equalcont{These authors contributed equally to this work.}
\author[9]{\fnm{Muntequa Imtiaz} \sur{Siraji}}\email{muntequaimtiaz@iut-dhaka.edu}
\equalcont{These authors contributed equally to this work.}
\author[10]{\fnm{Nafisa} \sur{Anjum}}\email{u1708003@student.cuet.ac.bd}
\equalcont{These authors contributed equally to this work.}
\author[11]{\fnm{Sumiya Rajjab} \sur{Shristy}}\email{sumiyashristy757@gmail.com}
\equalcont{These authors contributed equally to this work.}
\author[12]{\fnm{Tanvin} \sur{Rahman}}\email{tanvinrahman28@gmail.com}
\equalcont{These authors contributed equally to this work.}
\author[13]{\fnm{Mahmuda} \sur{Khatun}}\email{mahmudamohima14@gmail.com}
\equalcont{These authors contributed equally to this work.}
\author[14]{\fnm{Md Rubel} \sur{Dewan}}\email{mdrubeldewan99625@gmail.com}
\equalcont{These authors contributed equally to this work.}
\author[15]{\fnm{Mosaddeq} \sur{Hossain}}\email{tanjib.ju.42@gmail.com}
\equalcont{These authors contributed equally to this work.}
\author[16]{\fnm{Razia} \sur{Sultana}}\email{raziazuyenna@gmail.com}
\equalcont{These authors contributed equally to this work.}
\author[17]{\fnm{Ripel} \sur{Chakma}}\email{ripel.chakma13@gmail.com}
\equalcont{These authors contributed equally to this work.}
\author[18]{\fnm{Sonet Barua} \sur{Emon}}\email{sonetcste11@gmail.com}
\equalcont{These authors contributed equally to this work.}
\author[19]{\fnm{Towhidul} \sur{Islam}}\email{towhidul.islam@nub.ac.bd}
\equalcont{These authors contributed equally to this work.}
\author*[20]{\fnm{Mohammad Arafat} \sur{Hussain}}\email{mohammad.hussain@childrens.harvard.edu}
\cofirst{Co-first Authors.}

\affil[1]{\orgdiv{Department of Computer Science and Engineering}, \orgname{Ahsanullah University of Science and Technology}, \orgaddress{\city{Dhaka}, \postcode{1208}, \country{Bangladesh}}}

\affil[2]{\orgdiv{Department of Electrical and Telecommunication Engineering}, \orgname{Daffodil International University}, \orgaddress{\city{Dhaka}, \postcode{1216}, \country{Bangladesh}}}

\affil[3]{\orgdiv{Department of Industrial Engineering and Management}, \orgname{Khulna University of Engineering \& Technology (KUET)}, \orgaddress{\city{Khulna}, \postcode{9203}, \country{Bangladesh}}}

\affil[4]{\orgdiv{Physics and Biology Unit}, \orgname{Okinawa Institute of Science and Technology Graduate University (OIST)}, \orgaddress{\city{Okinawa}, \postcode{904-0412}, \country{Japan}}}

\affil[5]{\orgdiv{Department of Information and Communication Technology}, \orgname{Bangladesh University of Professionals}, \orgaddress{\city{Dhaka}, \postcode{1216}, \country{Bangladesh}}}

\affil[6]{\orgdiv{Department of Information and Communication Technology}, \orgname{Comilla University}, \orgaddress{\city{Cumilla}, \postcode{3506}, \country{Bangladesh}}}

\affil[7]{\orgdiv{Department of Biomedical Physics and Technology}, \orgname{University of Dhaka}, \orgaddress{\city{Dhaka}, \postcode{1000}, \country{Bangladesh}}}

\affil[8]{\orgdiv{Department of Computer Science and Engineering}, \orgname{International Islamic University Chittagong}, \orgaddress{\city{Chittagong}, \postcode{4318}, \country{Bangladesh}}}

\affil[9]{\orgdiv{Department of Electrical and Electronic Engineering}, \orgname{Islamic University of Technology}, \orgaddress{\city{Gazipur}, \postcode{1704},\country{Bangladesh}}}

\affil[10]{\orgdiv{Department of Electronics and Telecommunication Engineering}, \orgname{Chittagong University of Engineering and Technology}, \orgaddress{\city{Chittagong}, \postcode{4349}, \country{Bangladesh}}}

\affil[11]{\orgdiv{Department of Computer Science and Engineering}, \orgname{Notre Dame University Bangladesh}, \orgaddress{\city{Dhaka}, \postcode{1000}, \country{Bangladesh}}}

\affil[12]{\orgdiv{Department of Electrical and Electronic Engineering}, \orgname{Khulna University of Engineering \& Technology}, \orgaddress{\city{Khulna}, \postcode{9203}, \country{Bangladesh}}}

\affil[13]{\orgdiv{Department of Computer Science and Engineering}, \orgname{Institute of Science and Technology}, \orgaddress{\city{Dhaka}, \postcode{1209}, \country{Bangladesh}}}

\affil[14]{\orgdiv{Department of Electronics and Communication Engineering}, \orgname{Institute of Science and Technology}, \orgaddress{\city{Dhaka}, \postcode{1209}, \country{Bangladesh}}}

\affil[15]{\orgdiv{Department of Mathematics}, \orgname{Jahangirnagar University}, \orgaddress{\city{Dhaka}, \postcode{1342}, \country{Bangladesh}}}

\affil[16]{\orgdiv{Department of Electrical and Electronic Engineering}, \orgname{Daffodil International University}, \orgaddress{\city{Dhaka}, \postcode{1216}, \country{Bangladesh}}}

\affil[17]{\orgdiv{Department of Electrical and Electronic Engineering}, \orgname{Dhaka University of Engineering and Technology}, \orgaddress{\city{Gazipur}, \postcode{1707}, \country{Bangladesh}}}

\affil[18]{\orgdiv{Department of Computer Science and Telecommunication Engineering}, \orgname{Noakhali Science and Technology University}, \orgaddress{\city{Noakhali}, \postcode{3814}, \country{Bangladesh}}}

\affil[19]{\orgdiv{Department of Computer Science and Engineering}, \orgname{Northern University Bangladesh}, \orgaddress{\city{Dhaka}, \postcode{1230}, \country{Bangladesh}}}

\affil[20]{\orgdiv{Boston Children's Hospital}, \orgname{Harvard Medical School}, \orgaddress{\city{Boston}, \state{MA}, \postcode{02115}, \country{USA}}}

\abstract{The COVID-19 pandemic has affected millions of people globally, with respiratory organs being strongly affected in individuals with comorbidities. Medical imaging-based diagnosis and prognosis have become increasingly popular in clinical settings for detecting COVID-19 lung infections. Among various medical imaging modalities, ultrasound stands out as a low-cost, mobile, and radiation-safe imaging technology. In this comprehensive review, we focus on AI-driven studies utilizing lung ultrasound (LUS) for COVID-19 detection and analysis. We provide a detailed overview of both publicly available and private LUS datasets and categorize the AI studies according to the dataset they used. Additionally, we systematically analyzed and tabulated the studies across various dimensions, including data preprocessing methods, AI models, cross-validation techniques, and evaluation metrics. In total, we reviewed 60 articles, 41 of which utilized public datasets, while the remaining employed private data. Our findings suggest that ultrasound-based AI studies for COVID-19 detection have great potential for clinical use, especially for children and pregnant women. Our review also provides a useful summary for future researchers and clinicians who may be interested in the field.}


% \linenumbers

\maketitle

\section{Introduction}
\label{sec1}
\rd{The World Health Organization (WHO) declared Coronavirus Disease 2019 (COVID-19) a global pandemic in March 2020, and despite preventive measures, the virus has led to over 704 million cases and 7 million deaths worldwide~\citep{covid_worldometer}. COVID-19, like other respiratory infections, primarily affects the lungs, especially in individuals with comorbidities such as heart disease and diabetes~\citep{huang2020clinical,stokes2020coronavirus}. With the continued rise in cases and the emergence of new variants, medical imaging modalities such as computed tomography (CT), X-ray, and lung ultrasound (LUS) have become increasingly essential for diagnosing and monitoring COVID-19 lung infections~\citep{wang2022review,dong2020role,qian2020current}.}

Medical imaging is undeniably the most important tool for the diagnosis and management of treatments in clinical settings~\citep{willemink2020preparing}. Despite ultrasound being known to be a noisy imaging modality compared to various other imaging modalities with exceptional image quality (i.e., CT, magnetic resonance imaging, X-ray, etc.)~\citep{park2020non}, it stands out for being a low-cost, mobile, and, above all, non-ionizing medical imaging technology~\citep{yuan2021therapeutic}. Because ultrasound is radiation-safe, it is the preferred imaging modality for children and pregnant women~\citep{prentice2022use} and has been widely used in the detection and severity assessment of COVID-19 for the same patient group~\citep{allinovi2020lung}. Lung infection due to COVID-19 can be seen and assessed in LUS images.

\begin{figure}[!ht]
    \centering
    \includegraphics[width=3.75cm]{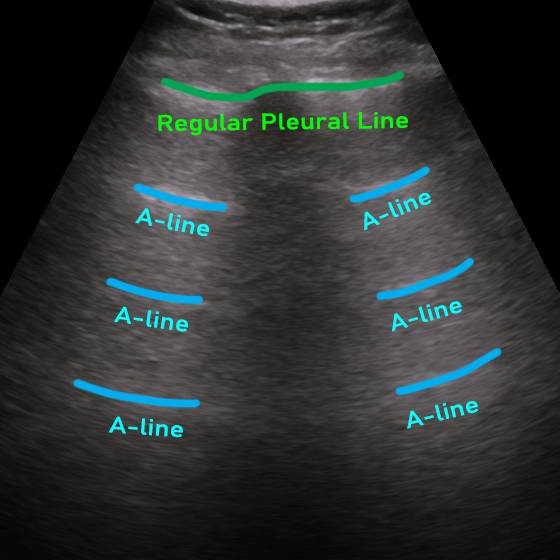}
    \includegraphics[width=3.75cm]{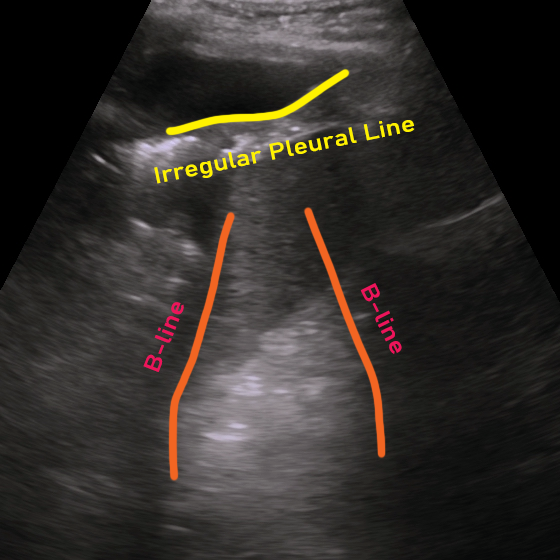}
    \includegraphics[width=3.75cm]{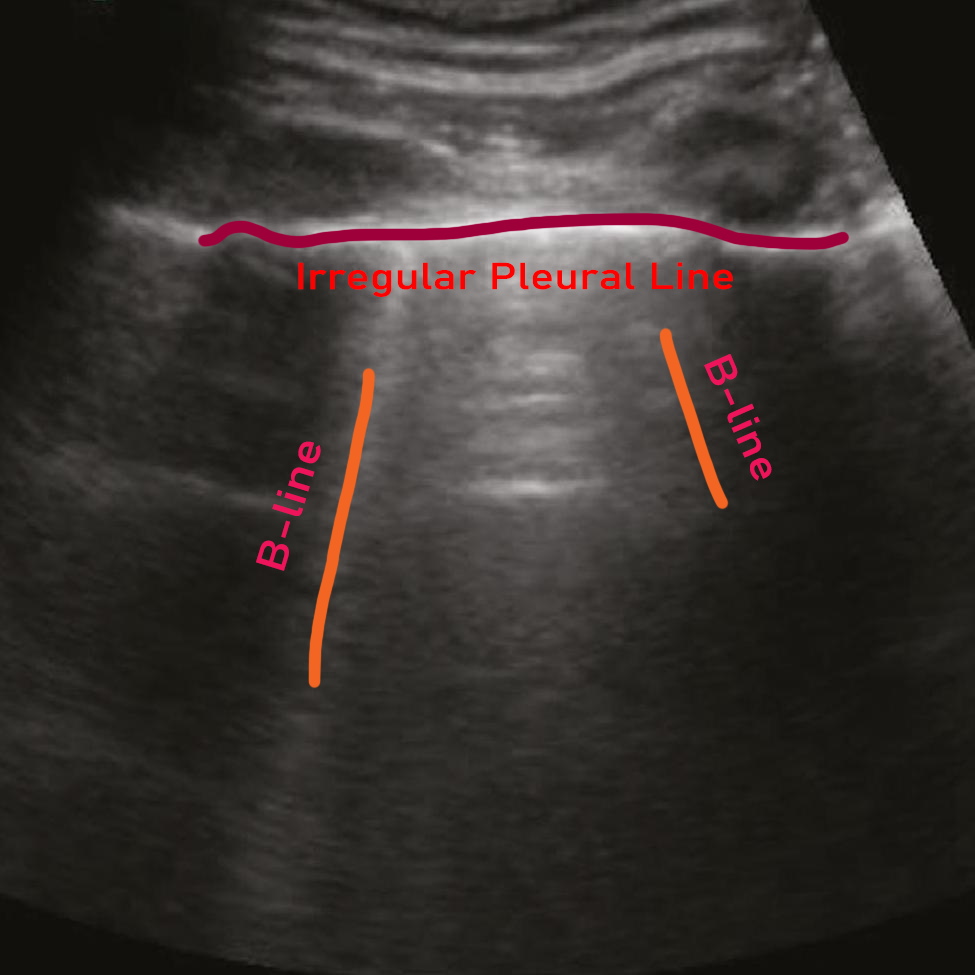}
    \caption{Demonstration of different types of lines that may appear in LUS images. A-lines are marked with blue, B-lines are marked with yellow, and the pleural line is marked with green~\citep{born2020pocovid}.}
    \label{US_lines}
\end{figure}

\begin{figure}[!ht]
    \centering
    \includegraphics[width=3.75cm]{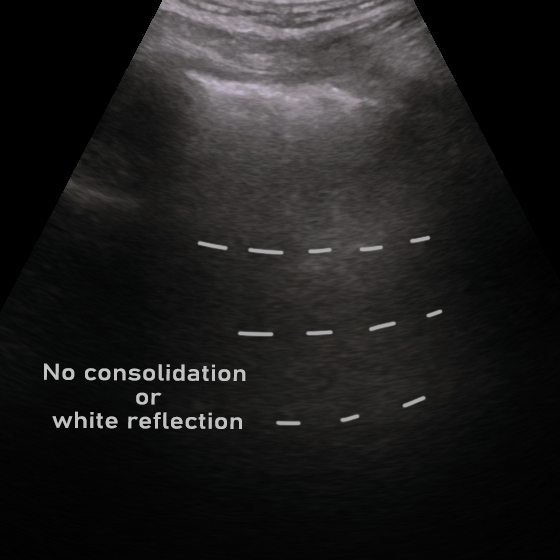}
    \includegraphics[width=3.75cm]{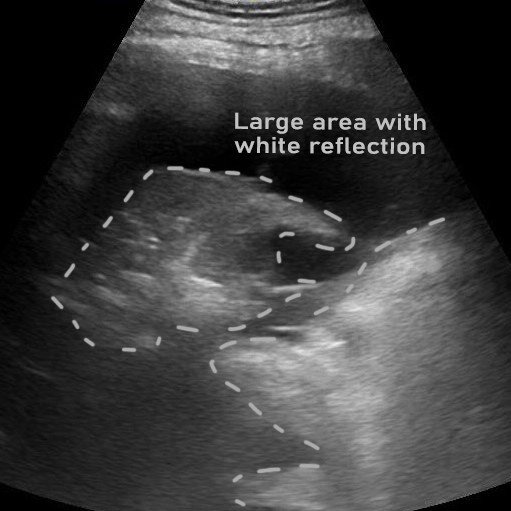}
    \includegraphics[width=3.75cm]{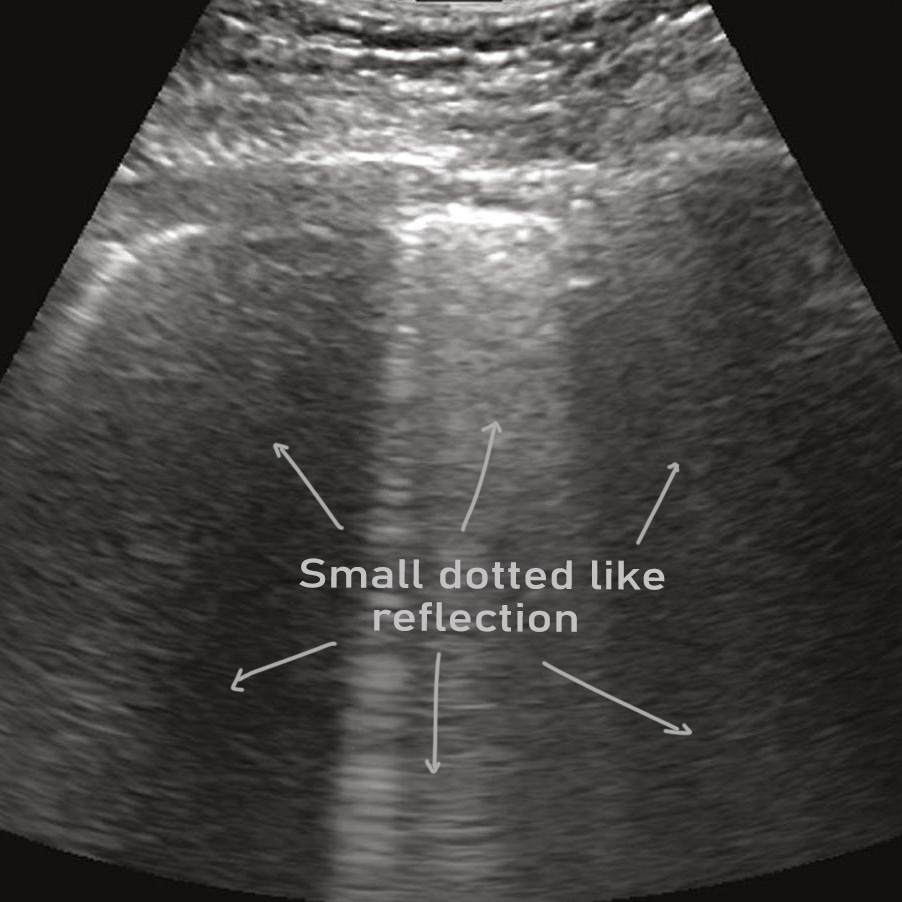}
    \caption{Example ultrasound images of a healthy lung (left), community-acquired pneumonia (CAP)-infected lung (middle), and COVID-19-infected lung (right)~\citep{born2020pocovid}.}
    \label{US_state}
\end{figure}

Typically, three major tasks can be performed on LUS images for COVID-19 patient management: (i) detection of pneumonia infection in the lung (e.g.,~\citep{born2020pocovid,born2021accelerating,diaz2019soft,gare2021dense}), (ii) pneumonia type/severity classification (e.g.,~\citep{carrer2020automatic,che2021multi,dastider2021integrated,frank2021integrating,roy2020deep}), and (iii) segmentation of infection in the lung (e.g.,~\citep{xue2021modality}). There are usually three types of artifacts that can appear in a LUS image, such as A-lines, B-lines, and irregular pleural lines (see Fig.~\ref{US_lines})~\citep{soldati2019role}. When ultrasound pulses reach the surface of the lung, healthy lungs exhibit horizontal lines parallel to the surface of the transducer, known as A-lines. On the other hand, a lung infected with pneumonia shows irregular pleural lines, as well as brightness in the lung (see Fig.~\ref{US_state}). On the contrary, COVID-19, a special kind of pneumonia, typically shows discreet vertical reverberation artifacts, known as B lines, which originate from the pleural surface (see Figs.~\ref{US_lines} and~\ref{US_state})~\citep{dastider2021integrated,frank2021integrating}. Based on the presence and appearance of these artifacts, pneumonia can be detected and classified as community-acquired pneumonia (CAP) and COVID-19, respectively. Finally, using segmentation, the spread of pneumonia can be estimated, which can be used for the severity scoring criteria for COVID-19~\citep{castelao2021findings, soldati2020proposal}.

To accelerate the detection and classification of CAP and COVID-19 in clinical settings, artificial intelligence (AI) algorithms~\citep{wang2022review,siddiqui2021early,mondal2021diagnosis,hussain2022active} have recently been introduced and have shown great promise, which may lessen the burden of expert radiologists/clinicians to detect and assess the severity of pneumonia. \rd{Several reviews have examined imaging modalities for COVID-19 detection using AI, but most have focused predominantly on CT and X-ray, with limited coverage of ultrasound-based methods. For instance,~\citet{ulhaq2020covid} reviewed 4,~\citet{alzubaidi2021role} 2, and~\citet{gudigar2021role} 3 ultrasound-based papers, with each identifying only 1 unique LUS dataset. Similarly,~\citet{huang2021artificial} and~\citet{siddiqui2021early} reviewed 1 and 3 papers, respectively, but identified no unique LUS datasets. Reviews by~\citet{liu2021review} and~\citet{mondal2021diagnosis} covered 2 papers each, identifying 1 and 0 unique datasets, respectively. Even the more recent reviews by~\citet{wang2022review,zhao2022review}, and~\citet{gursoy_overview_2023} reviewed 35, 9, and 5 papers on ultrasound methods, but identified only 3, 4, and 3 unique datasets, respectively. Lastly,~\citet{vafaeezadeh2024ultrasound} reviewed 2 papers that combined LUS with Vision Transformer (ViT) models yet identified no unique LUS datasets. In contrast, our study comprehensively reviews 60 LUS-specific papers, identifying 7 unique public and 21 private COVID-19 LUS datasets. Moreover, our review is organized around these datasets, allowing readers to focus on the methods employed for each specific dataset, which facilitates a more direct comparison of performance across studies. This structure enhances clarity and enables more informed conclusions regarding the effectiveness of different approaches in COVID-19 detection using LUS. In addition, this extensive dataset analysis fills a crucial gap left by previous reviews, which either focused on other modalities or reviewed a limited number of ultrasound papers with insufficient dataset exploration.}

\rd{Our study aims to address several key research questions (RQ) that have not been thoroughly explored in previous reviews on AI-based COVID-19 detection using LUS. Specifically, we explored the following questions: \textbf{RQ1:} What are the most commonly used public and private LUS datasets for COVID-19 detection? \textbf{RQ2:} How do the AI-based methods applied to COVID-19 ultrasound detection vary across different datasets, and what are the key performance metrics? \textbf{RQ3:} Which ultrasound image preprocessing and augmentation techniques are commonly used to enhance model performance? \textbf{RQ4:} What are the existing limitations and challenges in using ultrasound for COVID-19 detection, and how can future research address these challenges? 

In this comprehensive review, we address the above RQs to provide a thorough analysis of AI-based COVID-19 detection using LUS. Our contributions are as follows:
\begin{enumerate}
    \item We identify and catalog the most commonly used public and private LUS datasets for COVID-19 detection. 
    \item We analyze how different AI-based methods applied to COVID-19 ultrasound detection vary across these datasets and summarize the key performance metrics. 
    \item We review the ultrasound image preprocessing and augmentation techniques commonly employed to enhance model performance. 
    \item We identify and discuss the existing limitations and challenges in using ultrasound for COVID-19 detection and offer recommendations for addressing these challenges in future research. 
\end{enumerate}
}

\rd{\textbf{\textit{Search Strategy.}} We searched Google Scholar thoroughly for all scholarly publications: peer-reviewed journal papers, papers published in the proceedings of conferences or workshops, and non-peer-reviewed pre-prints from January 2020 to August 2024. Our search query was {\fontfamily{qcr}\selectfont (COVID-19 $\mid$ corona virus disease) (detect* $\mid$ predict* $\mid$ class*) (ultrasound)}. We applied a rigorous selection process to identify relevant articles for our review. The criteria for inclusion were: (1) the full text had to be accessible online or published in reputable journals or conferences indexed in databases such as PubMed, IEEE Xplore, Scopus, or Web of Science; (2) the article must have utilized AI techniques, either conventional machine learning or deep learning, specifically for the detection or analysis of COVID-19 using LUS data; (3) the hypothesis posed by the study had to be supported by robust qualitative and quantitative results; and (4) the article had to meet a minimum quality standard, ensuring no missing abstracts or methodologies, no reference errors, and clear figure legends and axis titles. Similar search strategies and selection criteria have been used in other recent reviews, e.g.,~\citep{azad2024advances}. In addition, we took great care to include all relevant studies utilizing LUS and AI for COVID-19 detection and analysis, though a few papers may have been inadvertently overlooked. Our goal, however, was to provide a comprehensive overview of the field. In total, we have reviewed 60 articles in this study.}

\rd{The remainder of the paper is organized as follows. Details of the datasets, the collection procedure of the ultrasound images, and the image processing techniques before the model building are presented in section~\ref{public_data}. An overview of the architecture of the AI models employed in the studies is presented in section~\ref{sec3}. Specific dataset-based studies with their methods and findings are tabulated and discussed in sections~\ref{sec4},~\ref{sec5},~\ref{sec6}, and~\ref{sec7}. Challenges, limitations, and gaps in the reviewed literature were summarized in section~\ref{sec8}. Discussion and potential future work are described in section~\ref{sec9}. Finally, concluding remarks are presented in section~\ref{con}.}

\section{\rd{Lung Ultrasound COVID-19 Datasets}}
\label{public_data}
Supervised learning using deep neural networks, a category of AI, has been extensively used for medical imaging applications in recent years~\citep{wynants2020prediction}. Adequate training of deep models for medical data requires prohibitive amounts of annotated data at the image/pixel/voxel level. Using such deep models on LUS data for COVID-19 detection and analysis is also not an exception. Furthermore, it is also critical to have public access to such datasets, as many research groups lack the clinical setup for data collection. In addition, reproducing a claimed performance by an AI method and possible future improvement greatly relies on access to the exact dataset. However, only a few publicly accessible LUS datasets are available. In this section, we discuss such datasets and their attributes.\\

\subsection{\rd{Publicly Accessible LUS COVID-19 Datasets}}
\label{subsec2.1}
\begin{table}[!ht]
\begin{center}
\begin{minipage}{350pt}
\tiny
\caption{\rd{List of publicly accessible LUS COVID-19 datasets.}}
\label{tab:public_data}
\begin{tabular}{|p{0.08cm} | p{2cm} | l | l | l | l |}
\hline
\textbf{Sl.} & \textbf{Dataset} & \textbf{Year} & \textbf{Number of Samples} & \textbf{Class Distribution} & \textbf{Note}\\
\hline
1 & Point-of-Care Ultrasound (POCUS) & 2020 & \makecell{(216 patients)\\202 videos\\59 images} & \makecell{COVID-19 (35\%)\\ Bacterial Pneumonia (28\%)\\Viral Pneumonia (2\%)\\Healthy (35\%)} & Link\footnote{\url{https://github.com/jannisborn/covid19_ultrasound/tree/master/data}}\\
\hline
2 & Italian COVID-19 Lung Ultrasound Database (ICLUS-DB) & 2020 & \makecell{(35 patients)\\277 videos\\58,924 frames}  & \makecell{Score 0: Continuous A-line (34\%)\\ Score 1: Alteration in A-line (24\%)\\ Score 2: Small consolidation (32\%)\\ Score 3: Large consolidation (10\%)} & Link\footnote{\url{https://www.disi.unitn.it/iclus}}\\
\hline
3 & COVIDx-US & 2021 & \makecell{242 videos\\29,651 images}  & \makecell{COVID-19 (29\%)\\CAP (20\%)\\Non-pneumonia diseases (39\%)\\Healthy (12\%)} & Link\footnote{\url{https://github.com/nrc-cnrc/COVID-US}}\\
\hline
4 & \rd{Boston Emergency Department Lung UltraSound (BEDLUS)} & 2022 & \makecell{(113 patients)\\1419 videos\\188,670 frames}  & \makecell{B-lines (50.7\%)\\No B-lines (49.3\%)} & Link\footnote{\url{https://doi.org/10.7910/DVN/GLCZRB}}\\
\hline
5 & \rd{COVID-19 Simulated and Labeled \textit{In Vivo} Dataset (CSLID)} & 2024 & \makecell{40,000 simulated images\\174 external \textit{in vivo} images\\958 internal \textit{in vivo} images}  & \makecell{A-line, B-line, and\\Consolidation features\\(10,000 phantoms per feature)} & Link\footnote{\url{https://gitlab.com/pulselab/covid19}}\\
\hline
6 & \rd{Fictional Lumen Dissection Dataset (FLDD)} & 2024 & \makecell{7050 ultrasound images}  & \makecell{Pneumonia (31\%)\\Normal (34\%)\\COVID (35\%)} & Link\footnote{\url{https://github.com/DASARINAGAVINOD/Covid-19_Ultrasound}}\\
\hline
7 & \rd{Lung Ultrasound COVID Phantom Dataset (LUCPD)} & 2024 & \makecell{564 phantom \\ultrasound images}  & \makecell{5 classes of object/artefact:\\ Rib, Pleural line,\\ A-line, B-line, \\and B-line confluence} & Link\footnote{\url{https://archive.researchdata.leeds.ac.uk/1263/}}\\
\hline
\end{tabular}
\end{minipage}
\end{center}
\end{table}

In Table~\ref{tab:public_data}, we list publicly accessible LUS COVID-19 datasets and their associated class labels. We briefly discuss each dataset below:\\

\noindent\textbf{Point-of-Care Ultrasound (POCUS):}~\citet{born2020pocovid,born2021accelerating} published and have been maintaining the POCUS dataset since 2020. This dataset initially contains a total of 261 lung ultrasound recordings by combining 202 videos and 59 still images collected from 216 patients. In this dataset, data from 92, 90, 73, and 6 are associated with COVID-19, healthy control, bacterial pneumonia, and viral pneumonia, respectively. These data were collected using either convex or linear probes. Each film in their dataset also comes with visual pattern-based expert annotation (e.g., B-Lines or consolidations).\\

\noindent\textbf{Italian COVID-19 Lung Ultrasound Database (ICLUS-DB):}~\citet{soldati2020proposal} published an internationally standardized acquisition protocol and four-level scoring schemes for lung ultrasound (LUS) in March 2020, shortly known as ICLUS-DB. This dataset contains 277 ultrasound videos (consisting of 58,924 frames) of 17 confirmed COVID-19, four suspected COVID-19, and 14 healthy subjects. 
These data were collected at various clinical centers in Italy using ultrasound scanners using either linear or convex probes. To evaluate the progress of pathology, this data consortium defined a four-level scoring system ranging from 0 to 3. The presence of continuous pleural-line and horizontal A-lines indicates a healthy lung with a score of 0. Score 1 is tagged for initial abnormality when alterations in the pleural line appear. Score 2 is more severe than one and is associated with small consolidations in the lung. Score 3 is the most severe grade, which is associated with the presence of a larger hyperechogenic area below the pleural surface (i.e., white lung).\\

\noindent\textbf{COVIDx-US:}~\citet{ebadi2022} published an open-access LUS benchmark dataset gathered from multiple sources in 2021. The dataset was assembled from a variety of sources (e.g., POCUS Atlas, GrepMed, Butterfly Network, and Life in the Fast Lane). This dataset (i.e., version 1.5) contains 242 videos (with 29,651 extracted images) corresponding to 71 COVID-19, 49 CAP, 94 non-pneumonia lung diseases, and 28 healthy classes.\\

\rd{
\noindent\textbf{Boston Emergency Department Lung UltraSound (BEDLUS):} The BEDLUS dataset consists of 1,419 LUS videos from 113 patients admitted to Brigham and Women's Hospital, MA, between November 2020 and March 2021 with flu-like symptoms \citep{lucassen2023deep}. The dataset includes 188,670 video frames, with 50.7\% of the videos positively labeled for B-lines, which are indicative of conditions like heart failure and pneumonia. Videos were acquired using a low-frequency transducer across various lung zones, with frames annotated for B-line origins by lung ultrasound experts. The dataset is preprocessed and de-identified, with available annotations and model parameters accessible online.\\

\noindent\textbf{COVID-19 Simulated and Labeled \textit{In Vivo} Dataset (CSLID):} \citet{zhao2024detection} utilized both simulated and in vivo ultrasound data to investigate B-line detection strategies in COVID-19 patients. Simulated data were generated using the MATLAB Ultrasound Toolbox, modeling a convex probe with 192 elements, a $73^{\circ}$ field of view, a 4 MHz center frequency, a 10 cm imaging depth, and a 60 MHz sampling frequency. This simulated data underwent standard ultrasound processing to create B-mode images with a 60 dB dynamic range.\\

\noindent\textbf{Fictional Lumen Dissection Dataset (FLDD):} \citet{vinod2024prognosis} accumulated approximately 7,050 ultrasound images from an ambiguous source(s), evenly distributed among COVID-19 positive cases, normal individuals, and pneumonia patients (2,350 images each). Images had been standardized to a resolution of 512$\times$512 pixels with Red-Green-Blue (RGB) reversion applied. The dataset was processed using gradient mapping. It also included computed Haralick features for both spatial (i.e., Gray-level difference matrix (GLDM), gray-level cooccurrence matrix (GLCM), and Texture) and frequency (i.e., Discrete wavelet transform (DWT), and fast Fourier transform (FFT)) domains. After modification of LUS data, \citet{vinod2024prognosis} termed their dataset as ``Fictional Lumen Dissection Dataset'' and made it public, intended for use in image segmentation and analysis tasks.\\

\noindent\textbf{Lung Ultrasound COVID Phantom Dataset (LUCPD):} This dataset~\citep{mclaughlan2024lung} consists of 564 phantom ultrasound images, focusing on five key objects and artifacts: Rib, Pleural line, A-line, B-line, and B-line confluence. These images were captured using a commercial lung ultrasound phantom (CAE Healthcare Inc., Blue Phantom COVID-19 Lung Ultrasound Simulator), which is designed to simulate features ranging from healthy to severely damaged lungs. The dataset was created by acquiring B-mode ultrasound videos with clinical systems and then extracting and annotating images using the VGG Image Annotator (VIA) tool. Multiple individuals with varying levels of ultrasound expertise labeled the images for segmentation purposes, ensuring high-quality annotations. This dataset provides a valuable resource for investigating LUS segmentation and learning to identify pathological signs associated with COVID-19.\\
}

\subsection{\rd{Non-Accessible LUS COVID-19 Private Datasets}}
\label{subsec2.2}
In contrast to the publicly accessible datasets described in section~\ref{subsec2.1}, some studies used private datasets, and some of these datasets are mentioned as available on request. However, these data sets have variations in terms of patient origin, hospital location, and data collection protocols. We list these datasets in Table~\ref{private_data} with the number of available samples and associated labels/classes. Below, we also briefly summarize the imaging protocols and types of transducers used in those datasets.

Regardless of the variation of ultrasound scanners, scanning areas on skin targeting the lung are typically similar across datasets. \citet{durrani2022automatic} considered six distinctive scanning regions in their study. \citet{panicker2021approach} adopted the scan protocol of~\citet{soldati2020proposal} and also aimed at six acquisition points for data extraction. \citet{quentin2020extracting} scanned on ten thoracic sites in their study. Although video of the costophrenic region was excluded in~\citep{arntfield2020development}, most studies followed a twelve-zone scanning protocol for the data acquisition process~\citep{camacho2022artificial, chen2021quantitative, huang2022evaluation, la2021deep, wang2021semi, xue2021modality}. Furthermore, \citet{mento2021deep} used fourteen scanning areas, following the scan protocol of by~\citet{soldati2020proposal}. Another study~\citep{roshankhah2021investigating} followed the scan protocol by~\citet{mento2021impact} and \citet{perrone2021new}.

Variations in transducer types and frequency were also observed in the studies. For example, some studies used low-frequency (1–5 MHz) curved array~\citep{chen2021quantitative,huang2022evaluation,camacho2022artificial,panicker2021approach} and phased array~\citep{durrani2022automatic,arntfield2020development} transducers. On the other hand, \citet{roshankhah2021investigating} used both linear and convex transducers in multi-sites with a wide range of center frequencies. Similarly, \citet{la2021deep} used both linear and convex transducers with a frequency of 5 and 12 MHz, respectively, and \citet{mento2021deep} used 3.5 to 6.6 MHz in their study. \rd{\cite{kuroda2023artificial} used a dataset that includes 56 subjects, with 41 COVID-19 patients and 15 controls. Lung point-of-care ultrasound (POCUS) and CT scans were performed, analyzing 397 lung zones in patients and 180 in controls, with some zones excluded due to positioning limitations. The retrospective study conducted by \cite{sagreiya2023automated} involved data from multiple institutions and public databases, and a total of 52 ultrasound scans were performed using various scanners. Different probes were used depending on the imaging requirements, with findings verified by radiologists experienced in LUS related to COVID-19. 

\citet{faita2024covid} collected 2,067 LUS videos from 135 COVID-19-positive patients across two cohorts: 1,564 videos from 104 patients in 2020 (cohort 1) and 503 videos from 31 patients in 2022 (cohort 2). The videos were annotated with a clinically validated severity score by expert sonographers, ranging from 0 to 3 based on symptom severity. \citet{kimura2024effectiveness} used a private dataset that included video images from 69 patients with suspected congestive heart failure (CHF), captured using Lumify devices with a 3 MHz cardiac transducer. \citet{li2024knowledge} used the dataset, including 1,447 frames, categorized into 113 mild, 21 moderate, eight severe, and 25 critical cases. The dataset typically contains eight images per examination, with imaging settings tailored to each patient.

}

\rd{

\tiny
\renewcommand\thefootnote{\alph{footnote}} 
\begin{longtable}{| p{0.1cm} | p{2cm} | p{0.4cm} | p{1.8cm} | p{1cm} | p{2.5cm} | p{0.9cm} |}
\caption{List of private (publicly non-accessible) COVID-19 ultrasound datasets. Acronyms- N: number of samples, Tr: training, Va: validation, and Te: test.}
\label{private_data}\\
\hline
\textbf{Sl.} & \textbf{Dataset} & \textbf{Year} & \textbf{N} & \textbf{Tr/Va/Te} & \textbf{Classes} & \textbf{Note} \\
\hline
\endfirsthead  % header material

\caption*{\textbf{Table 2 (Continued):} List of private (non-accessible publicly) COVID-19 ultrasound datasets. Acronyms- N: number of samples, Tr: training, Va: validation, and Te: test.}\\
\hline
\textbf{Sl.} & \textbf{Dataset} & \textbf{Year} & \textbf{N} & \textbf{Tr/Va/Te} & \textbf{Classes} & \textbf{Note} \\
\hline
\endhead  % header material

1 & London Health Sciences Centre's two tertiary hospitals (Canada) \citep{arntfield2020development} & 2020 & (243 patients); 600 videos; 121,381 frames & $\sim$80/20 & COVID, Non-COVID, Hydrostatic Pulmonary Edema & - \\
\hline
2 & ULTRACOV (Ultrasound in Coronavirus disease)~\citep{camacho2022artificial} & 2022 & (28 COVID-19 patients) 3 sec video each & - & A-Lines, B-Lines, consolidations, and pleural effusions & Available upon request \\
\hline
3 & Huoshenshan Hospital (Wuhan, China)~\citep{chen2021quantitative} & 2021 & (31 patients) 1,527 images & - & Normal, septal syndrome, interstitial-alveolar syndrome, white lung & \href{https://bio-hsi.ecnu.edu.cn/}{Source Link}\footnote{https://bio-hsi.ecnu.edu.cn/} \\
\hline
4 & Royal Melbourne Hospital (Australia)~\citep{durrani2022automatic} & 2022 & (9 patients) 27 videos; 3,827 frames & - & Normal, consolidation/collapse & Available upon request \\
\hline
5 & Ultrasound lung data \citep{ebadi2022} & 2021 & (300 patients) 1530 videos; 287,549 frames & 80/20 & A-line artifacts, B-line artifacts, presence of consolidation/pleural effusion & - \\
\hline
6 & Huoshenshan Hospital (Wuhan, China)~\citep{huang2022evaluation} & 2022 & (31 patients); 2,062 images & - & Normal, septal syndrome, interstitial-alveolar syndrome, white lung & \href{https://bio-hsi.ecnu.edu.cn/}{Source Link}\footnote{https://bio-hsi.ecnu.edu.cn/} \\
\hline
7 & Fondazione IRCCS Policlinico San Matteo's Emergency Department (Pavia, Italy)~\citep{la2021deep} & 2021 & (450 patients) 2,908 frames & 75/15/10 & A-lines with two B-lines, slightly irregular pleural line, artifacts in 50\% of the pleura, damaged pleural line, visible consolidated areas, damaged pleura/irregular tissue & - \\
\hline
8 & Third People's Hospital of Shenzhen (China)~\citep{liu2020semi} & 2020 & (71 COVID-19 patients) 678 videos; 6,836 images & - & A-line, B-line, pleural lesion, pleural effusion & - \\
\hline
9 & Fondazione Policlinico Universitario Agostino Gemelli (Rome, Italy), Fondazione Policlinico San Matteo (Pavia, Italy) \citep{mento2021deep} & 2021 & (82 patients) 1,488 videos; 314,879 frames & - & 4 severity levels \citep{soldati2020proposal} & - \\
\hline
10 & CHUV (Lausanne, Switzerland) \citep{quentin2020extracting} & 2020 & (193 patients) 1,265 videos; 3,455 images & 80/20 & True (experts' approval), False (experts' disapproval) & - \\
\hline
11 & Various online sources \citep{nabalamba2022machine} & 2022 & 792 images & - & COVID-19, healthy & - \\
\hline
12 & Spain, India \citep{panicker2021approach} & 2021 & (10 subjects) 400 videos, 5,000 images & - & A-lines, lack of A-lines, the appearance of B-lines, the confluent appearance of B-lines, the appearance of C-lines & Available upon request \\
\hline
13 & Private clinics (Lima, Peru) \citep{rojas2021detection} & 2021 & 1,500 images & - & Healthy, COVID-19 & Available upon request \\
\hline
14 & BresciaMed (Brescia, Italy), Valle del Serchio General Hospital (Lucca, Italy), Fondazione Policlinico Universitario A. Gemelli IRCCS (Rome, Italy), Fondazione Policlinico Universitario San Matteo IRCCS (Pavia, Italy), and Tione General Hospital (Tione, Italy) \citep{roshankhah2021investigating} & 2021 & (32 patients) 203 videos; 1,863 frames & 90/10 & Healthy, indentation of pleural line, discontinuity of the pleural line, white lung & - \\
\hline
15 & Beijing Ditan Hospital (Beijing, China)~\citep{wang2021semi} & 2021 & (27 COVID-19 patients) 13 moderate, 7 severe, 7 critical & - & Severe, non-severe & - \\
\hline
16 & Cancer Center of Union Hospital, West of Union Hospital, Jianghan Cabin Hospital, Jingkai Cabin Hospital, Leishenshan Hospital \citep{xue2021modality} & 2021 & (313 COVID-19 patients) 10-second video from each & - & Normal, presence of 3-5 B-lines, $\geq$6 B-lines or irregular pleura line, fused B-lines or thickening pleura line, consolidation & - \\
\hline
17 & Juntendo University Graduate School of Medicine (Tokyo, Japan) \citep{kuroda2023artificial} & 2023 & (56 subjects) 577 lung zones & - & Count of B-lines in each zone & - \\
\hline
18 & Collection of Data from Various Unspecified Institutions and Public Databases \citep{sagreiya2023automated} & 2023 & 52 ultrasound examinations & - &  & - \\
\hline
19 & Italian National Research Council and University of Pisa (Italy) \citep{faita2024covid} & 2024 & 104 patients 1564 LUS videos & - & 3 severity levels & - \\
\hline
20 & Mayo Clinic, Rochester, Minnesota \citep{kimura2024effectiveness} & 2024 & 110 video clips 69 patients & - & Normal, Mild-moderately abnormal, Severely abnormal & - \\
\hline
21 & Beijing Ditan Hospital (Beijing, China) \citep{li2024knowledge} & 2024 & (152 patients) 1,447 frames & - & COVID severity levels: Mild, Moderate, Severe, Critical & - \\
\hline
\end{longtable}
% \end{landscape}

}
\normalsize

\begin{figure}[!ht]
    \centering
    \includegraphics[width=0.9\linewidth]{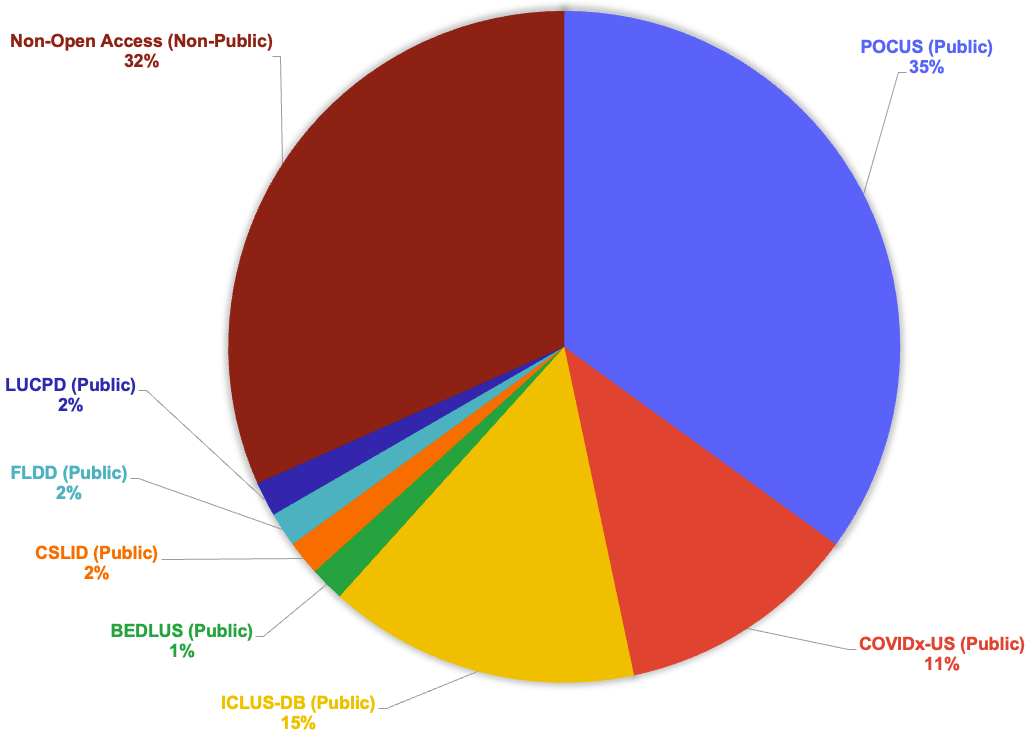}
    \caption{\rd{A pie-chart showing the percentage of reviewed articles in this study per LUS datasets.}}
    \label{piechart}
\end{figure}

\rd{Figure~\ref{piechart} presents a pie-chart showing the distribution of datasets of the reviewed articles in this study, with POCUS accounting for 21 (35\%), Non-open Access for 19 (32\%), ICLUS-DB for 9 (15\%), COVIDx-US for 7 (11\%), and other sparsely used public datasets for 4 (7\%).}

\subsection{Data Pre-processing and Augmentation}
Various image processing techniques are typically used before feeding the data to AI models. Image processing techniques include, but are not limited to, curve-to-linear conversion, image resizing, intensity normalization, standardization, augmentation, etc. In this section, we briefly discuss different image pre-processing techniques used in the reviewed articles. 

\subsubsection{Curve-to-linear\rd{/Linear-to-curve} Conversion}\label{curve_to_linear}
Acquired ultrasound videos and images using convex transducers are typically fan-shaped (i.e., narrower close to the probe surface while wider at depth). In contrast, ultrasound videos and images that use linear transducers are usually rectangular. Thus, harmonizing images acquired by convex and linear transducers requires the conversion of fan-shaped images to rectangular images \rd{and vice versa}. Therefore, various automatic built-in conversion techniques in the scanner, as well as external user-defined interpolation techniques~\citep{bottenus2015acoustic}, are typically used for this conversion task, and ultrasound-based COVID-19 AI studies are not an exception, e.g.,~\citep{carrer2020automatic,li2024knowledge, zeng2024covid}. 

\subsubsection{Image Resizing}\label{subsubsec2.1.1}
Image resizing is the most common image pre-processing technique used for AI model training. Typically, ultrasound images come with various resolutions in terms of pixel count. On the other hand, AI models, especially deep learning models, typically require all input images to be of equal dimension. In addition, the larger input image dimension and the number of channels cause a higher computational overhead in the AI model optimization process. Therefore, AI studies often resize input images to a widely used common dimension across datasets. Most of the reviewed articles in this paper, for example,~\citep{ebadi2021automated, rojas2021detection, nabalamba2022machine,quentin2020extracting,karar2021lightweight,karnes2021adaptive,perera2021pocformer,song2023covid, madhu2024xcovnet}, etc., also used the common image dimension of 224$\times$224 pixels as well-known computer vision deep learning models are typically designed to intake images of 224$\times$224 pixels. However, other image dimensions are also found for ultrasound COVID-19 studies. For example, \citet{karar2021adversarial} resized all ultrasound images to 28$\times$28 pixels to avoid a higher computational overhead. In addition, \citet{nehary2023lung}, \citet{mateu2022covid}, \citet{lucassen2023deep}, \citet{durrani2022automatic}, \citet{vinod2024prognosis}, \citet{muhammad2021covid}, and \citet{gare2021dense} resized their ultrasound images to 128$\times$128, 254$\times$254, 384$\times$256, 806$\times$550, 512$\times$512, and 624$\times$464 pixels, respectively.

\subsubsection{Intensity Normalization}\label{subsubsec2.1.2}
Intensity normalization is another common image pre-processing technique used in AI studies. This process ensures a common intensity range across images and datasets. In most cases, all image data are converted to a common intensity range of [0, 1], or [0, 255]~\citep{perera2021pocformer}, followed by mean subtraction and division by standard deviation~\citep{muhammad2021covid,adedigba2021deep,quentin2020extracting,roshankhah2021investigating,wang2021semi,lucassen2023deep, li2024knowledge, torti2024gpu}.

\subsubsection{Image Augmentation}\label{subsubsec2.1.3}
Image augmentation is a widely used technique in AI studies, which is used to increase the amount of training data and the variation and diversity in the appearance of an image. one of the most prevalent steps that have been executed in most of the studies. Various conventional (as in~\citet{hussain2017segmentation}) and learning-based data augmentation~\citep{momeny2021learning} techniques are present in the literature. Conventional image augmentation techniques such as image cropping, random rotation, horizontal and vertical flipping, histogram equalization, random image shifting, zooming in and out, and/or a combination of these operations, etc., are more prevalent in AI studies, and articles in this review (e.g., \citet{born2020pocovid,gare2021dense,muhammad2021covid,roy2020deep,arntfield2020development,adedigba2021deep,la2021deep,nabalamba2022machine,rojas2021detection,khan2023benchmark,howell2024deep,faita2024covid,song2023covid,zhao2024detection}) mostly adopted this type of augmentation. 

\subsubsection{Other Image Processing Techniques}\label{subsubsec2.1.4}
Apart from the common image pre-processing techniques discussed above, other processes are often used in ultrasound AI studies. Ultrasound images are known to be a noisy modality ~\citep{pal2021review}. Therefore, ultrasound-based studies often use noise reduction filters for pre-processing of images~\citep{dastider2021integrated}, such as circular averaging filter~\citep{de2016smoothing}, median filter~\citep{hussain2012direct}, non-linear diffusion filter~\citep{hussain2015towards}, contrast-limited adaptive histogram equalization (CLAHE)~\citep{sadik2021}, etc. 

\citet{ebadi2022} performed several pre-processing operations to make resulting ultrasound images in COVIDx-US\footnote{https://github.com/nrc-cnrc/COVID-US} dataset easily usable to AI models. They cropped video frames into rectangular windows to remove the background or visible text from the image periphery. Any video frame with a moving pointer on it was also ignored when frames were extracted to use as images.

\rd{Other image preprocessing techniques used by reviewed articles in this study include image blurring~\citep{khan2023benchmark,zhao2024detection}, elastic warping~\citep{khan2023benchmark}, and variable time-gain compensation~\citep{howell2024deep}.}

\section{\rd{AI in LUS COVID-19 Studies}}
The accuracy of identifying COVID-19 infection and assessing its severity is based primarily on the expertise of clinicians,  which is often difficult and time-consuming. To overcome this limitation, AI approaches have been widely used in recent years. AI approaches used in COVID-19 ultrasound studies can be categorized into conventional machine learning (CML) and deep learning (DL) approaches. CML approaches (e.g., support vector machine (SVM), linear regression, etc.) typically require hand-engineering of features, which are often difficult to define optimally~\citep{hussain2017segmentation}. Overcoming this limitation, DL using convolutional neural networks (CNN) has exploded in popularity throughout the last decade. Various CNN architectures have been widely used on natural image and medical image-based classification and segmentation tasks. However, medical imaging data are often very difficult to collect, which results in a small training data cohort. To overcome this limitation, DL on medical imaging often leverages the transfer learning strategy, where the deep model is pre-trained on a much larger natural image dataset and then finetuned on the target smaller medical data. This transfer learning strategy is also used in many articles (e.g.,~\citet{diaz2021,al2021covid,barros2021,nabalamba2022machine,rojas2021detection}) we reviewed in this study. In addition, many studies in this review (e.g.,~\citet{diaz2021,born2020pocovid}) used cross-validation techniques to avoid overfitting. 

\subsection{AI Models}
\label{sec3}
\begin{figure}[!ht]
    \centering
    \includegraphics[width=12cm]{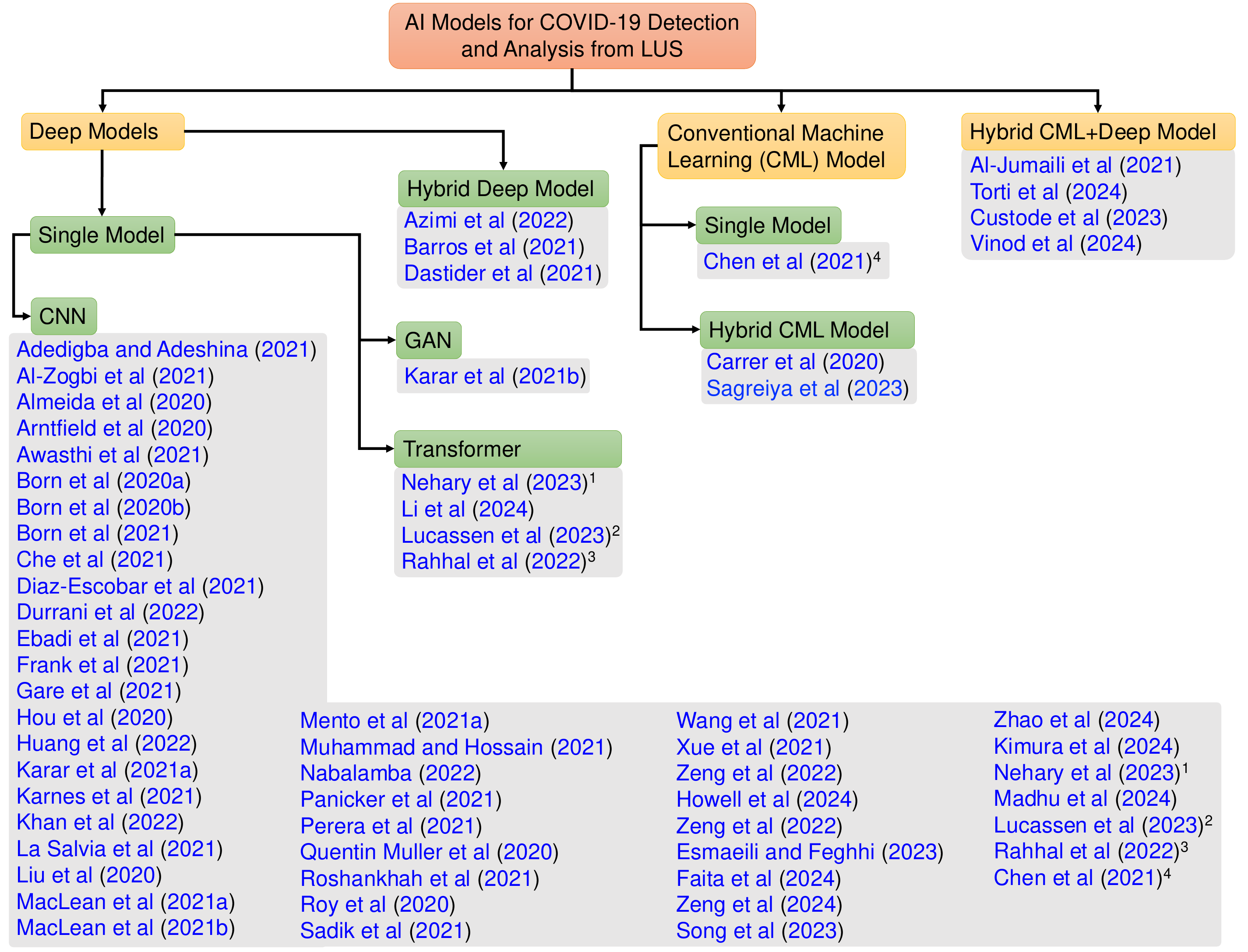}
    \caption{\rd{Organization of reviewed articles in terms of the AI model types and configurations. Hybrid models represent those studies that used two or more types of machine learning strategies together as part of a common predictive model. Four studies used both CML and DL strategies, but they were not part of a common predictive model. Those articles are shown twice under different categories and indicated with common superscripts, 1-4.}}
    \label{flowchart}
\end{figure}

In Table~\ref{tab:ai_models}, we list all the articles we reviewed in this study and the corresponding AI methods used by those articles. We also mark in the table whether a study used CML, DL, or both. We see in the table that only \rd{three studies used CML approaches (see rows 12, 40, and 44 of Table~\ref{tab:ai_models}), and five studies combined CML and DL (see rows 2, 14, 53, 55, and 56 of Table~\ref{tab:ai_models})}. Except for these studies, all other studies we reviewed used DL approaches. This tendency to prefer DL approaches over CML approaches is motivated by the fact that DL models are capable of learning optimal feature representation by themselves without requiring manual intervention and the availability of more complex and powerful computation facilities. In Fig.~\ref{flowchart}, we organized all the reviewed articles in terms of the type of AI model and configuration. We also describe different types of AI models used by state-of-the-art LUS COVID-19 studies in the following sections.
\begin{table}[!ht]
\begin{center}
\tiny
\begin{minipage}{330pt}
\caption{A list of the articles reviewed in this study and the corresponding AI methods used by those articles. Acronyms- Sl.: serial, CML: conventional machine learning, DL: deep learning, RNN: recurrent neural network, SVM: support vector machine, LSTM: long short-term memory, STN: spatial transformer network, GAN: generative adversarial network, ViT: vision transformer, gMLP: multi-layer perceptron with gating.}
\label{tab:ai_models}
\begin{adjustbox}{max width=\textwidth}
\begin{tabular}{| l | l | l | c | c |}

%\toprule
\hline
\textbf{Sl.} & \textbf{Studies} & \textbf{AI Methods} & \textbf{CML} & \textbf{DL}\\
\hline 
 
1	&	\citet{adedigba2021deep} & SqueezeNet, MobileNetV2 &\xmark&\checkmark\\\hline
2	&	\citet{al2021covid} & ResNet-18, RestNet-50, NASNetMobile, GoogleNet, SVM &\checkmark&\checkmark\\\hline
3	&	\citet{alzogbi2021} & DenseNet&\xmark&\checkmark\\\hline
4	&	\citet{almeida2020} & MobileNet&\xmark&\checkmark\\\hline
5	&	\citet{arntfield2020development} & Xception&\xmark&\checkmark\\\hline
6	&	\citet{awasthi2021} & MiniCOVIDNet&\xmark&\checkmark\\\hline
7	&	\citet{azimi2022covid} & InceptionV3, RNN&\xmark&\checkmark\\\hline
8	&	\citet{barros2021} & Xception-LSTM&\xmark&\checkmark\\\hline
9	&	\citet{born2020pocovid} & VGG-16&\xmark&\checkmark\\\hline
10	&	\citet{DBLP:journals/corr/abs-2009-06116} & VGG-16&\xmark&\checkmark\\\hline
11	&	\citet{born2021accelerating} & VGG-16&\xmark&\checkmark\\\hline
12	&	\citet{carrer2020automatic} & Hidden Markov Model, Viterbi Algorithm, SVM&\checkmark&\xmark\\\hline
13	&	\citet{che2021multi} & Multi-scale Residual CNN&\xmark&\checkmark\\\hline
14	&	\citet{chen2021quantitative} & 2-layer NN, SVM, Decision Tree&\checkmark&\checkmark\\\hline
15	&	\citet{diaz2021}& InceptionV3, VGG-19, ResNet-50, Xception&\xmark&\checkmark\\\hline
16	&	\citet{dastider2021integrated}& Autoencoder-based Hybrid CNN-LSTM&\xmark&\checkmark\\\hline
17	&	\citet{durrani2022automatic}& Reg-STN&\xmark&\checkmark\\\hline
18	&	\citet{ebadi2021automated}& Kinetics-I3D&\xmark&\checkmark\\\hline
19	&	\citet{frank2021integrating}& ResNet-18, MobileNetV2, DeepLabV3++&\xmark&\checkmark\\\hline
20	&	\citet{gare2021dense}& Reverse Transfer Learning on U-Net&\xmark&\checkmark\\\hline
21	&	\citet{hou2020}& Saab transform-based SSL, CNN &\xmark&\checkmark\\\hline
22	&	\citet{huang2022evaluation}& Non-local channel attention ResNet &\xmark&\checkmark\\\hline
23	&	\citet{karar2021lightweight}& MobileNet, ShuffleNet, MENet, MnasNet&\xmark&\checkmark\\\hline
24	&	\citet{karar2021adversarial}& A semi-supervised GAN, a modified AC-GAN&\xmark&\checkmark\\\hline
25	&	\citet{karnes2021adaptive}& Few-shot learning using MobileNet&\xmark&\checkmark\\\hline
26	&	\citet{khan2022deep}& CNN&\xmark&\checkmark\\\hline
27	&	\citet{la2021deep}& ResNet-18, ResNet-50&\xmark&\checkmark\\\hline
28	&	\citet{liu2020semi}& Multi-symptom multi-label (MSML) network &\xmark&\checkmark\\\hline
29	&	\citet{maclean2021covid}& COVID-Net US&\xmark&\checkmark\\\hline
30	&	\citet{maclean2021initial}& ResNet&\xmark&\checkmark\\\hline
31	&	\citet{mento2021deep}& STN, U-Net, DeepLabV3+&\xmark&\checkmark\\\hline
32	&	\citet{muhammad2021covid}& CNN&\xmark&\checkmark\\\hline
33	&	\citet{nabalamba2022machine}& VGG-16, VGG-19, ResNet&\xmark&\checkmark\\\hline
34	&	\citet{panicker2021approach}& LUSNet (a U-Net like network for ultrasound images)&\xmark&\checkmark\\\hline
35	&	\citet{perera2021pocformer}& \rd{ViT}&\xmark&\checkmark\\\hline
36	&	\citet{quentin2020extracting}& ResNet-18&\xmark&\checkmark\\\hline
37	&	\citet{roshankhah2021investigating}& U-Net&\xmark&\checkmark\\\hline
38	&	\citet{roy2020deep}& STN, U-Net, U-Net++, DeepLabV3, Model Genesis&\xmark&\checkmark\\\hline
39	&	\citet{sadik2021}& DenseNet-201, ResNet-152V2, Xception, VGG-19, NasNetMobile&\xmark&\checkmark\\\hline
40	&	\citet{wang2021semi}& SVM&\checkmark&\xmark\\\hline
41	&	\citet{xue2021modality}& U-Net&\xmark&\checkmark\\\hline
42	&	\citet{zeng2022}& COVID-Net US-X&\xmark&\checkmark\\\hline
\rd{43}	&	\citet{kuroda2023artificial}& AI-POCUS (Model specifics are not disclosed) &-&-\\\hline
\rd{44}	&	\citet{sagreiya2023automated}& Calculated Lung Ultrasound (CLU)&\checkmark&\xmark\\\hline
\rd{45}	&	\citet{khan2023benchmark}& ResNet-18, ResNet-50, ResNet-101, DenseNet-121, DenseNet-201, InceptionV3, RegNetX, EfficientNetB7&\xmark&\checkmark\\\hline
\rd{46}	&	\citet{howell2024deep}& Lightweight U-Net&\xmark&\checkmark\\\hline
\rd{47}	&	\citet{esmaeili2023covid}& Uniform Local Binary Pattern on Five Intersecting Planes and CNN (ULBPFP-Net)&\xmark&\checkmark\\\hline
\rd{48}	&	\citet{faita2024covid}& Inflated 3D Convolutional Network (I3D)&\xmark&\checkmark\\\hline
\rd{49}	&	\citet{zeng2024covid}& COVID-Net L2C-ULTRA&\xmark&\checkmark\\\hline
\rd{50}	&	\citet{song2023covid}& COVID-Net USPro&\xmark&\checkmark\\\hline
\rd{51}	&	\citet{zhao2024detection}& U-Net&\xmark&\checkmark\\\hline
\rd{52}	&	\citet{kimura2024effectiveness}& CNN&\xmark&\checkmark\\\hline
\rd{53}	&	\citet{torti2024gpu}& ResNet-50$+$K-means&\checkmark&\checkmark\\\hline
\rd{54}	&	\citet{nehary2023lung}& VGG16, ViT&\xmark&\checkmark\\\hline
\rd{55}	&	\citet{custode2023multi}& STN, U-Net+DeepLabV3+, Decision Tree&\checkmark&\checkmark\\\hline
\rd{56}	&	\citet{vinod2024prognosis}& GAN+Random Forest&\checkmark&\checkmark\\\hline
\rd{57}	&	\citet{madhu2024xcovnet}& Xception Convolutional Neural Network (XCovNet)&\xmark&\checkmark\\\hline
\rd{58}	&	\citet{li2024knowledge}& Knowledge Fusion with Latent Representation (KFLR) Transformer&\xmark&\checkmark\\\hline
\rd{59}	&	\citet{lucassen2023deep}& ResNet3D-18, ResNet(2+1)D-18, 3D U-Net, DenseNet-121, EfficientNetB0, ViT, DeepLabV3+ &\xmark&\checkmark\\\hline
\rd{60}	&	\citet{rahhal2022contrasting}& EfficientNetB2, ViT, gMLP &\xmark&\checkmark\\\hline
\end{tabular}
\end{adjustbox}
\end{minipage}
\end{center}
\end{table}

\subsubsection{Convolutional Neural Networks (CNN)}
\rd{\textit{SqueezeNet, MobileNetV2, VGG-16/19, NasNetMobile, DenseNet-121/201, ResNet-18/50/101/152V2, ResNet(2+1)D-18, ResNet3D-18, InceptionV3, GoogleNet, EfficientNetB0/B2/B7, XCovNet:}} These CNNs are DL models specifically designed for image processing tasks. They typically consist of convolutional layers that extract features from input images and pooling layers that downsample the feature maps. These models typically include building blocks such as convolutional filters, activation functions (e.g., ReLU), pooling layers (e.g., MaxPooling), and fully connected layers for classification. Convolution layers apply a set of learnable filters to the input data. Each filter performs a convolution operation, which involves sliding across the input and computing dot products with local regions. This process helps extract important visual features, such as edges, textures, and patterns. Pooling layers, on the other hand, downsample the spatial dimensions of the feature maps generated by convolutional layers. They reduce the computational complexity and extract the most important information by summarizing groups of values in the feature maps. The most common type of pooling is max pooling, which selects the maximum value within each pooling region. Furthermore, fully connected layers connect every neuron from the previous layer to the subsequent layer, similar to traditional neural networks. Fully connected layers are responsible for the final classification or regression task, combining the extracted features to make predictions. Most of the articles we reviewed in this study used different types of CNNs, for example, SqueezeNet~\citep{adedigba2021deep}, MobileNetV2~\citep{adedigba2021deep}, VGG-16~\citep{born2020pocovid,nabalamba2022machine,DBLP:journals/corr/abs-2009-06116,nehary2023lung}, VGG-19~\citep{nabalamba2022machine,sadik2021}, ResNet-18~\citep{al2021covid,quentin2020extracting,la2021deep,khan2023benchmark}, ResNet-50~\citep{al2021covid,la2021deep,khan2023benchmark,torti2024gpu}, \rd{ResNet-101~\citep{khan2023benchmark}}, ResNet-152V2~\citep{sadik2021}, \rd{DenseNet-121~\citep{khan2023benchmark,lucassen2023deep}, DenseNet-201~\citep{sadik2021,khan2023benchmark}}, InceptionV3~\citep{azimi2022covid,khan2023benchmark}, GoogleNet~\citep{al2021covid}, \rd{EfficientNetB0~\citep{lucassen2023deep}, EfficientNetB2~\citep{rahhal2022contrasting}, EfficientNetB7~\citep{khan2023benchmark}, XCovNet~\citep{madhu2024xcovnet}} and NasNetMobile~\citep{al2021covid,sadik2021}.

\rd{\textit{ULBPFP-Net:} It is a DL model that combines texture analysis with CNNs for image classification. It uses the Uniform Local Binary Pattern on Five Intersecting Planes (ULBPFP), a texture descriptor that captures local patterns from multiple image planes, to extract robust features. These features are then processed by a CNN, which learns complex patterns and representations for accurate classification. The integration of ULBP with CNNs enhances the ability of a model ability to capture both texture and spatial information, improving performance in tasks like medical imaging. \citet{esmaeili2023covid} used this approach in their study.

\textit{Inflated 3D Convolutional Network (I3D):} I3D is a deep learning model designed for video analysis, extending 2D convolutional networks by inflating 2D filters into 3D. This allows the model to capture both spatial and temporal features from video data, making it highly effective in processing and classifying videos, particularly for tasks like action recognition. A study by \citet{faita2024covid} utilized the I3D approach in analyzing ultrasound COVID-19 data. \textit{Kinetics-I3D}, on the other hand, builds upon the I3D model by combining it with the Kinetics dataset, a large-scale video dataset specifically designed for action recognition. Kinetics-I3D leverages pre-training on this dataset, enabling the model to learn rich spatiotemporal representations from millions of labeled video clips, effectively capturing motion and temporal dependencies. One study~\citep{ebadi2021automated} we reviewed employed Kinetics-I3D in their study.}

\subsubsection{Recurrent Neural Networks (RNN)}
RNNs are a type of neural network that can process sequential data by capturing temporal dependencies. They are commonly used for tasks involving sequential inputs or outputs, such as natural language processing and time series analysis. RNNs have recurrent connections that allow information to flow from one time step to the next. This enables the network to maintain a memory of previous inputs and utilize that information to make predictions or analyze the current input. At each time step, an RNN produces an output based on the current input and the hidden state from the previous time step. The hidden state serves as the network's memory, storing information about previous inputs. It is updated and passed along to the next time step, allowing the network to learn and capture long-term dependencies in the sequence. RNNs can be ``unfolded'' in time, creating a series of interconnected layers that correspond to each time step. This unfolding helps visualize the flow of information through the network and enables the application of backpropagation through time, a training algorithm that adjusts the network's weights based on the sequence of inputs and desired outputs. One of the articles we reviewed in this study used RNN~\citep{azimi2022covid}.

\subsubsection{COVID-Net}
\textit{COVID-Net US, COVID-Net US-X, \rd{COVID-Net USPro, COVID-Net L2C-ULTRA}:} These architectures are specifically developed for the detection and diagnosis of COVID-19 from medical imaging, particularly chest X-ray images. COVID-Net US is a CNN architecture designed for the classification of chest X-ray images to detect COVID-19 cases. It has been trained on a large dataset of X-ray images and is capable of distinguishing COVID-19 from other respiratory conditions. The architecture of COVID-Net US includes convolutional layers for feature extraction, pooling layers for downsampling, and fully connected layers for classification. COVID-Net US-X is an extended version of COVID-Net US with improved performance and capabilities. It incorporates advancements such as additional layers, refined architecture, or enhanced training techniques to enhance the accuracy and reliability of COVID-19 detection from chest X-ray images. Two articles we reviewed in this study used COVID-Net US~\citep{maclean2021covid} and COVID-Net US-X~\citep{zeng2022}. \rd{On the other hand, COVID-Net USPro is a few-shot learning model designed to classify unlabelled data by comparing it to labeled examples. It operates by creating a prototype representation for each class from labeled data and assigns unlabelled data to the class with the closest prototype in an embedding space, using a distance metric to measure similarity. This model is trained in an episodic setting, allowing it to handle limited labeled data effectively. One of the studies~\citep{song2023covid} in our review used COVID-Net USPro. Another study~\citep{zeng2024covid} in our review used COVID-Net L2C-ULTRA, which is a data augmentation learning method designed to address data scarcity and heterogeneity in POCUS images. It enhances the diversity of training data by applying random projective and piecewise affine transformations, helping linear probe images resemble those from convex probes. This technique improves the generalization of deep models by exposing them to more diverse and visually consistent data.}

\textit{MiniCOVIDNet:} It is a compact and efficient neural network architecture designed for COVID-19 detection from chest X-ray images. It is specifically developed to provide a smaller model that can be deployed on resource-constrained devices or in scenarios where computational efficiency is important. The architecture of MiniCOVIDNet typically includes convolutional layers, pooling layers, and fully connected layers, aiming to accurately classify X-ray images as COVID-19 positive or negative while minimizing computational requirements. One of the articles we reviewed in this study used MiniCOVIDNet~\citep{awasthi2021}.

\subsubsection{Long Short-Term Memory (LSTM)}
LSTM is a type of RNN architecture that addresses the vanishing gradient problem of traditional RNNs and is capable of capturing long-term dependencies in sequential data. LSTMs are widely used in various tasks involving sequential data, such as natural language processing, speech recognition, and time series analysis. The key feature of LSTM networks is their memory cell, which allows them to retain information over long sequences and selectively forget or update that information. LSTMs achieve this through a set of gates, including an input gate, a forget gate, and an output gate. These gates regulate the flow of information and enable the LSTM to remember or forget specific information based on the context. One of the articles we reviewed in this study used LSTM~\citep{dastider2021integrated}.

\textit{Xception-LSTM:} Xception-LSTM refers to a specific model architecture that combines the Xception CNN with an LSTM layer. Xception is a deep CNN architecture that was proposed as an extension of the Inception architecture. It introduces a novel concept called \rd{``depthwise separable convolutions''} to reduce the number of parameters and computations required by traditional convolutions. Another article we reviewed in this study used Xception-LSTM~\citep{barros2021}.

\subsubsection{Hidden Markov Model (HMM)}
HMM and Viterbi Algorithm are both fundamental concepts in the field of probabilistic modeling and sequential data analysis. One of the articles we reviewed in this study used both HMM and Viterbi Algorithm~\citep{carrer2020automatic}.

\textit{Hidden Markov Model:} An HMM is a statistical model that represents a system with unobservable (hidden) states and observable outputs. It is a generative model that assumes the underlying system can be modeled as a Markov process, where the current state depends only on the previous state. However, the actual state is not directly observable; instead, it emits observable symbols or outputs. HMMs have been widely used in various applications such as speech recognition, natural language processing, bioinformatics, and pattern recognition. 

\textit{Viterbi Algorithm:} The Viterbi Algorithm, on the other hand, is an efficient dynamic programming algorithm used to find the most likely sequence of hidden states in a Hidden Markov Model. Given a sequence of observations, the Viterbi Algorithm computes the optimal sequence of hidden states that maximizes the probability of the observations. It takes into account both the transition probabilities between states and the emission probabilities of observations from the states. The algorithm iteratively computes the most likely path by considering the accumulated probabilities at each time step, resulting in the most probable sequence of hidden states.

\subsubsection{Generative Adversarial Networks (GAN)}
GAN is a class of machine learning models that consists of two neural networks, namely the generator and the discriminator, which are trained together in a competitive setting. The generator network takes random noise as input and generates synthetic samples, such as images, based on that noise. The objective of the generator is to generate samples that resemble real data as closely as possible. On the other hand, the discriminator network takes both real samples from the dataset and synthetic samples from the generator as input and aims to classify them correctly as real or fake. The discriminator's objective is to distinguish between real and generated samples accurately. During training, the generator and discriminator are trained in alternating steps. The generator tries to fool the discriminator by generating realistic samples, while the discriminator aims to improve its ability to distinguish real from fake samples. This back-and-forth training process creates a competitive dynamic where the generator improves its ability to generate realistic samples, and the discriminator becomes more adept at discriminating between real and fake samples.

\textit{AC-GAN:} Auxiliary Classifier GAN (AC-GAN) is an extension of the GAN framework that includes an auxiliary classifier in addition to the discriminator. The auxiliary classifier is a separate network that is trained to predict additional class labels or attributes associated with the generated samples. This helps in controlling the generated samples to have specific attributes or belong to specific classes. The addition of the auxiliary classifier in AC-GAN allows for more control over the generated samples and enables the generation of samples conditioned on specific attributes or classes. It has been used in various applications, including image synthesis, text-to-image generation, and image-to-image translation, where the generation process can be guided by specific attributes or class labels. One of the articles we reviewed in this study used a semi-supervised GAN and AC-GAN~\citep{karar2021adversarial}.

\rd{\subsubsection{Transformer}
Transformer is a DL architecture designed for sequence modeling tasks like natural language processing. It relies on self-attention mechanisms to capture relationships between all elements in a sequence, allowing it to process data in parallel and learn long-range dependencies more efficiently than recurrent models.

\textit{Vision Transformer (ViT):} ViT applies the Transformer architecture to image data by dividing an image into patches, treating each patch as a token similar to words in a sentence. It processes these patches through self-attention mechanisms, allowing the model to capture global image features and achieve strong performance in image classification tasks. Some studies~\citep{perera2021pocformer,nehary2023lung,lucassen2023deep,rahhal2022contrasting} we reviewed in this work used ViT for LUS COVID-19 data analysis and classification.

\textit{Knowledge Fusion with Latent Representation (KFLR) Transformer:} KFLR Transformer is a specialized Transformer model that integrates multiple sources of information by learning a shared latent representation. This fusion of knowledge from different domains or modalities improves the ability of a model to handle complex tasks, as it combines diverse insights while maintaining high-quality latent feature representations. One study in our review used KFLR Transformer to predict the severity of COVID-19 from LUS~\citep{vinod2024prognosis}.}

\subsubsection{Spatial Transformer Network (STN)}
STN is a type of neural network module that can be integrated into deep learning architectures to enable the spatial transformation of input data. The purpose of the STN network is to learn spatial transformations, such as rotations, translations, scaling, and cropping, that can be applied to input images or feature maps. The key idea behind the STN network is to introduce a spatial transformer module that can learn to automatically align and transform input data to improve the overall performance of the model. The module consists of three main components- (i) Localization Network: The localization network takes the input data and learns to predict the parameters of the spatial transformation. It typically consists of convolutional and fully connected layers that extract features and output the transformation parameters, such as translation, rotation, and scaling. (ii) Grid Generator: The grid generator takes the predicted transformation parameters from the localization network and generates a set of sampling grid points. These grid points define how the input data should be transformed to align with the desired output. (iii) Sampler: The sampler takes the input data and the generated grid points and performs the spatial transformation. It applies interpolation techniques, such as bilinear interpolation, to sample the input data at the grid points and produce the transformed output. By incorporating the STN network into a larger neural network architecture, the model can learn to automatically adjust and align the input data to improve performance. The STN module can be trained end-to-end with the rest of the network using backpropagation, allowing the model to learn the appropriate spatial transformations for the given task. Several articles we reviewed in this study used Transformer or STN~\citep{perera2021pocformer,roy2020deep,mento2021deep}.

\textit{Reg-STN:} Reg-STN stands for Regression Spatial Transformer Network. It is an extension of the STN that incorporates regression-based localization instead of classification-based localization. One of the articles we reviewed in this study used Reg-STN~\citep{durrani2022automatic}.

\subsubsection{U-Net}
U-Net is a convolutional neural network architecture that was specifically designed for biomedical image segmentation but has since been applied to various other domains. It consists of an encoder-decoder structure with skip connections. The encoder part gradually reduces the spatial dimensions while capturing hierarchical features, and the decoder part upsamples the feature maps and recovers the spatial resolution. The skip connections help preserve fine-grained details by concatenating feature maps from the encoder to the corresponding decoder layers. U-Net has been widely used for tasks such as medical image segmentation, cell segmentation, and more. Several articles we reviewed in this study used U-Net\citep{mento2021deep,roshankhah2021investigating,roy2020deep,xue2021modality,gare2021dense}.

\textit{U-Net++:} U-Net++ is an extension of the U-Net architecture that aims to further enhance the segmentation performance. It introduces a nested and densely connected skip pathway structure. In U-Net++, each encoder block is connected to all corresponding decoder blocks through skip connections, creating a more extensive and interconnected network. This architecture allows for better information flow and feature reuse across different scales, leading to improved segmentation accuracy and boundary delineation. One of the articles we reviewed in this study used Reg-UNet++~\citep{roy2020deep}.

\textit{LUSNet:} LUSNet (Lung Ultrasound Net) is a specific implementation of the U-Net-like network architecture designed for lung ultrasound image segmentation. It incorporates the U-Net framework with modifications tailored for lung ultrasound images. LUSNet leverages the inherent characteristics of lung ultrasound images, such as the presence of artifacts, pleural lines, and specific structures like A-lines and B-lines, to perform accurate segmentation. By adopting the U-Net architecture to the unique properties of lung ultrasound images, LUSNet aims to provide reliable segmentation for various lung-related applications, including disease diagnosis and monitoring. One of the articles we reviewed in this study used LUSNet~\citep{panicker2021approach}.

\subsubsection{Few-shot Learning}
Few-shot learning is a machine learning paradigm that addresses the problem of learning from limited labeled data. In traditional machine learning approaches, a large amount of labeled data is typically required to train a model effectively. However, in real-world scenarios, collecting and annotating large datasets can be time-consuming, expensive, or impractical. Few-shot learning aims to overcome this limitation by enabling models to learn new concepts or tasks with only a few labeled examples. It focuses on the ability of a model to generalize and adapt to new classes or tasks based on a small amount of labeled data, often referred to as the ``support set.'' The key idea in few-shot learning is to leverage prior knowledge or information learned from related tasks or classes to facilitate learning on new tasks or classes with limited examples. This is achieved through various techniques such as meta-learning, where the model learns to quickly adapt to new tasks based on its previous experience, or by using generative models to synthesize additional training examples. One of the articles we reviewed in this study used Few-shot learning~\citep{karnes2021adaptive}.

\subsubsection{Transfer Learning}
Transfer learning is a machine learning technique that involves leveraging knowledge learned from one task or domain to improve performance on another related task or domain. In transfer learning, a pre-trained model that has been trained on a large dataset and a related task is used as a starting point for a new task. The idea is that the pre-trained model has learned general features and representations that can be useful for the new task, even if the specific classes or labels are different. By using transfer learning, the model can benefit from the knowledge and representations learned from the large pre-training dataset, which can save training time and improve performance, especially when the target dataset is limited or the target task is challenging.

\textit{Reverse Transfer Learning:} Reverse transfer learning, on the other hand, is a less commonly used term and refers to the process of transferring knowledge or models from a target domain or task back to the source domain or task. It involves utilizing the information or insights gained during the target task and applying them to improve the performance of the original source model or task. In reverse transfer learning, the experience and knowledge gained from the target domain or task are used to enhance the performance or update the source model. This can be particularly useful when the target task provides new insights or data that can benefit the original source task or when there is a feedback loop between the target and source tasks. One of the articles we reviewed in this study used Reverse transfer learning~\citep{gare2021dense}.

\subsubsection{Support Vector Machine (SVM)}
SVM is a popular CML algorithm used for classification and regression tasks. SVM is known for its ability to handle both linear and non-linear data by finding an optimal hyperplane that separates different classes or predicts the continuous target variable. In the case of classification, SVM aims to find the best decision boundary that maximally separates different classes in the input feature space. This decision boundary is determined by a subset of training samples called support vectors. SVM works by mapping the input data into a higher-dimensional feature space using a kernel function, which allows the algorithm to find a hyperplane that effectively separates the classes. The choice of the kernel function, such as linear, polynomial, or radial basis function (RBF), affects the SVM's ability to handle complex patterns and non-linear relationships. In the case of regression, SVM aims to find a hyperplane that best fits the data while minimizing the error between the predicted and actual target values. The SVM regression algorithm aims to find a balance between fitting the data closely and controlling the complexity of the model to avoid overfitting. Several articles we reviewed in this study used SVM~\citep{al2021covid,carrer2020automatic,chen2021quantitative,wang2021semi}.

\subsubsection{Decision Tree}
\label{dt}
A decision tree is another supervised CML algorithm that is commonly used for classification and regression tasks. It is a flowchart-like structure where internal nodes represent feature tests, branches represent the outcomes of those tests, and leaf nodes represent the predicted class or value. The decision tree algorithm recursively splits the data based on different features to create a tree-like model that can make predictions. At each internal node, a decision is made based on the values of a particular feature, and the data is split into subsets accordingly. This splitting process continues until a stopping criterion is met, such as reaching a maximum tree depth or a minimum number of samples at a node. During training, the decision tree algorithm determines the optimal splits by evaluating different feature and split point combinations based on certain criteria, such as Gini impurity or information gain. The goal is to create splits that result in homogeneous subsets with respect to the target variable. Once the decision tree is trained, it can be used to make predictions by traversing down the tree based on the feature values of an unseen sample. The path followed through the tree leads to a leaf node, which provides the predicted class for classification tasks or the predicted value for regression tasks. Decision trees are popular due to their interpretability and simplicity. They can handle both categorical and numerical features and can capture non-linear relationships between features and the target variable. However, decision trees are prone to overfitting, especially when the tree becomes too complex. Techniques like pruning and ensemble methods, such as random forests, are often used to mitigate overfitting and improve the performance of decision trees. \rd{Two articles we reviewed in this study used Decision Tree~\citep{chen2021quantitative,custode2023multi}.}

\rd{\subsubsection{K-means}
K-means is an unsupervised clustering algorithm that partitions data into a specified number of clusters ($k$). It assigns each data point to the nearest cluster by minimizing the distance to the cluster’s centroid, which is iteratively updated until the assignments stabilize. It is commonly used for unsupervised learning tasks like grouping similar data points or segmenting datasets. One of the studies~\citep{torti2024gpu} we reviewed used K-means on the generated features by ResNet-50.

\subsubsection{Random Forest}
Random Forest is a CML ensemble method used for both classification and regression tasks. It builds multiple decision trees (described in section~\ref{dt}) during training and combines their predictions to improve accuracy and reduce overfitting. Each tree is trained on a random subset of the data, making Random Forest robust and capable of handling complex data patterns. One of the studies~\citep{torti2024gpu} we reviewed used Random Forest on the generated data by GAN.}

\subsubsection{Other Architectures}
\textit{Saab transform-based successive subspace learning model:} It refers to a specific approach for feature extraction and dimensionality reduction in image processing and computer vision tasks. It is based on a series of transformations called the Successive Subspace Learning (SSL) framework, with the Saab transform being one of the key components. The Saab transform is a non-linear transformation that aims to capture discriminative and compact representations of image features. It operates on local image patches and applies a series of operations, including patch-wise mean removal, PCA (Principal Component Analysis), and non-linear transformation using sigmoid functions. These operations are performed successively to obtain a hierarchical representation of the input image.

\textit{Non-local channel attention ResNet:} It refers to a variant or modification of the ResNet architecture that incorporates non-local channel attention mechanisms. It aims to enhance the representation power of ResNet models by introducing non-local operations that capture long-range dependencies across channels. In the context of the ``Non-local channel attention ResNet,'' the term ``non-local'' refers to the inclusion of non-local operations within the ResNet architecture. Non-local operations are designed to model relationships between spatial or temporal positions in an image or sequence, capturing dependencies that extend beyond local regions. The ``channel attention'' component refers to the specific attention mechanism applied to the channels or feature maps within the ResNet. Attention mechanisms help the network focus on relevant information by assigning importance weights to different channels or feature maps. This allows the model to selectively emphasize or suppress certain channels during feature extraction, enhancing its ability to capture important patterns or details. One of the articles we reviewed in this study used Non-local channel attention ResNet~\citep{huang2022evaluation}.

\textit{The Multi-symptom multi-label (MSML) network:} MSML network refers to a neural network architecture designed to address the problem of multi-label classification in the context of medical diagnosis. In medical diagnosis, patients may exhibit multiple symptoms simultaneously, and it is important to accurately predict the presence or absence of various medical conditions or diseases based on these symptoms. The MSML network is specifically designed to handle this scenario by taking into account multiple symptoms as inputs and predicting multiple labels as outputs. It is a type of deep learning architecture that combines techniques from multi-label classification and DNNs. The MSML network typically consists of multiple layers of interconnected neurons, including input layers, hidden layers, and output layers. The input layer receives the symptom information as input features, which are then processed through the hidden layers to extract relevant patterns and representations. Finally, the output layer produces predictions for multiple medical conditions or diseases associated with the given symptoms. One of the articles we reviewed in this study used MSML network~\citep{liu2020semi}.

\textit{Model Genesis:} ``Model Genesis'' refers to a specific deep learning architecture designed for medical image analysis tasks. It is an automated framework that aims to generate high-quality medical image segmentation models using a combination of deep convolutional neural networks (CNNs) and genetic algorithms. The concept behind Model Genesis is to leverage the power of genetic algorithms to automatically search and optimize the architecture and hyperparameters of a CNN for a given medical image segmentation task. The genetic algorithm explores a population of potential CNN architectures, evaluates their performance on a specific objective function (such as accuracy or Dice coefficient), and evolves the population over successive generations to improve the model's performance. The genetic algorithm in Model Genesis operates by using principles inspired by natural selection and evolution. It applies genetic operators such as mutation, crossover, and selection to create new CNN architectures from existing ones, gradually improving their fitness and convergence toward an optimal solution. By integrating genetic algorithms into the architecture search process, Model Genesis aims to automate the design of CNN models for medical image segmentation tasks, reducing the need for manual architecture design and hyperparameter tuning. This approach can potentially save time and effort for researchers and practitioners in the medical imaging field, allowing them to focus more on the analysis and interpretation of medical images. One of the articles we reviewed in this study used Model Genesis~\citep{roy2020deep}.

\subsection{Loss Functions}\label{subsec3.5}
A classification model can be defined as $\hat{y}=f_\theta(x)$, where the AI model $f_\theta$ is parameterized by a set of parameters $\theta$ and an input image $x$ is assigned to the most probable class $\hat{y}$. Given a training set of ultrasound images $x_i$ and their ground truth class $y_i \{(x_i,y_i); i = 1,..., N\}$, training a classification model consists of finding the model parameters $\theta$ that minimize loss $\mathcal{L}$, such as:
\begin{equation}
    \theta^*=\arg \min_{\theta}\sum_{i=1}^N\mathcal{L}(\hat{y}_i\mid y_i)
\end{equation}
Therefore, the choice of the appropriate loss function $\mathcal{L}$ is important, and we briefly discuss the loss functions used in the articles reviewed in this study.

\subsubsection{Cross-entropy Loss}\label{subsubsec3.5.2}
Training an AI model on a binary decision-making task (e.g., COVID-19 \textit{vs.} CAP, or COVID-19 \textit{vs.} healthy, etc.) usually utilizes binary cross-entropy or simply cross-entropy loss defined as:
\begin{equation}
    \mathcal{L}_{CE}(X,Y;\theta)=-\frac{1}{N}\sum_{i=1}^{N}{y_i}\times log(\hat{y}_i)+(1-y_i)\times log(1-\hat{y}_i).
\end{equation}
The cross-entropy loss appears in the majority of ultrasound COVID-19 AI studies (e.g., \citet{born2020pocovid,born2021accelerating,diaz2021,gare2021dense,perera2021pocformer,che2021multi,frank2021integrating}).

\subsubsection{Categorical Cross-entropy}\label{subsubsec3.5.3}
Categorical cross-entropy works on multiclass (more than two classes; e.g., COVID-19 \textit{vs.} CAP \textit{vs.} Healthy) classification problems. This loss is typically used in an AI model when the model must select one or more categories among numerous possible categories/classes. This loss can be defined as:
\begin{equation}
    \mathcal{L}_{CCE}(X,Y;\theta)=-\frac{1}{N}\sum_{i=1}^{N} y_i\times log(\hat{y}_i).
\end{equation}
Like cross-entropy loss, categorical cross-entropy loss also appears in many ultrasound COVID-19 AI studies (e.g., \citet{barros2021,karar2021lightweight,sadik2021}).

\subsubsection{L1 Loss}\label{l1_loss}
L1 loss, also known as mean absolute loss, is typically used when an AI model is tasked to predict a continuous value (e.g., the distance between two landmarks, optimal location for lung scanning using ultrasound, etc.). It is defined as:
\begin{equation}
    \mathcal{L}_{1}(X,Y;\theta)=\sum_{i=1}^{N}{\mid y_{true}-y_{predict}\mid},
\end{equation}
where $y_{true}$ and $y_{predict}$ are the ground truth and predicted continuous values, respectively.~\citet{alzogbi2021} used this loss function to train their deep model to predict landmarks for optimal ultrasound scanning.

\subsubsection{Focal Loss}\label{focal_loss}
The focal loss is a dynamically scaled cross-entropy loss and is used when there is a class in the training data. Focal loss incorporates a modulating term in the conventional cross-entropy loss so that it can emphasize learning from difficult data samples that lead to misclassification more often. This loss is defined as:
\begin{equation}
    \label{focal_l1}
    \mathcal{L}_{FL}(X,Y;\theta)=-\frac{1}{N}\sum_{i=1}^{N}(1-\hat{y}_i)^\gamma \times log(\hat{y}_i),
\end{equation}
where $\gamma$ controls the weight of different samples and $\gamma=0$ transforms Eq.~\ref{focal_l1} into a binary cross-entropy loss. \citet{awasthi2021} used focal loss in their ultrasound-based COVID-19 study.

\subsubsection{Soft Ordinal (SORD) Loss}\label{sord}
When output classes are independent of each other, their relative order in the loss calculation during deep model training does not matter. This scenario allows using one-hot encoding, i.e., setting all wrong classes to be infinitely far from the true class. However, there exists a soft order among classes in an ordinal regression scenario, where certain categories are more correct than others with respect to the true label \citep{diaz2019soft} (i.e., a true class is no longer infinitely far from false classes, resulting in a continuity among classes). For these continuously related classes, \citet{roy2020deep} introduced a modified cross-entropy, called soft ordinal (SORD) loss, defined as:
\begin{equation}
    \label{focal_l}
    \mathcal{L}_{SORD}(X,Y;\theta)=-\sum_{i=1}^{\mid\mathcal{N}\mid} \left(\frac{e^{-\delta(n,i)}}{\sum_{j\in\mathcal{N}}e^{-\delta(j,i)}}\right) \times log\left(\frac{e^{f_\theta(x_i)}}{\sum_j^{\mid\mathcal{N}\mid}e^{f_\theta(x_j)}}\right),
\end{equation}
where $\mid\mathcal{N}\mid$ is the set of possible soft-valued classes, $n$ is a possible ground truth soft value, $\delta$ is a user-defined distance (e.g., weighted square distance) between scores/levels, $f_\theta$ is the deep model, and $x_i$ is the $i$-th input data.

\rd{
\subsubsection{Dice Loss}
Dice Loss is a widely used loss function for image segmentation tasks, especially in medical imaging. It is derived from the Dice Similarity Coefficient (DSC), which measures the overlap between the predicted segmentation and the ground truth. The Dice coefficient ranges from 0 to 1, where 1 represents perfect overlap, and 0 indicates no overlap at all. The Dice Loss is formulated to minimize this overlap difference, making it ideal for handling class imbalance in segmentation problems. Dice Loss is defined as:

\begin{equation}
    \label{dice_loss}
    \mathcal{L}_{Dice} = 1 - \frac{2 \cdot \lvert A \cap B \rvert}{\lvert A \rvert + \lvert B \rvert}
\end{equation}

Here, $A$ represents the set of predicted pixels, and $B$ represents the set of ground truth pixels. The intersection $\lvert A\cap B\rvert$ measures the common pixels between the prediction and the ground truth, while $\lvert A \rvert$ and $\lvert B \rvert$ represent the sizes of the predicted and ground truth sets, respectively. Dice loss effectively addresses the issue of class imbalance, which is common in segmentation tasks where the background class may dominate the image. By focusing on maximizing the overlap between predicted and true regions, Dice loss ensures that even small, underrepresented regions are accurately captured by the model. This makes it particularly effective in medical applications, such as tumor or organ segmentation, where precise boundary delineation is essential. Its ability to handle imbalanced data and focus on correct segmentation overlap makes it a preferred choice for segmentation tasks in DL.~\citet{zhao2024detection} and ~\citet{vinod2024prognosis} used Dice loss in their LUS COVID-19 study.
}

\subsection{Evaluation Criteria}\label{subsec3.6}
The effectiveness of a predictive model is measured by an evaluation metric. This often entails building a model on a dataset, testing it on a holdout dataset that was not used during training, and comparing the predictions to the holdout dataset's expected values. The metrics compare the actual class label to the predicted class label for the classification problems. The different studies reviewed in this article used different types of evaluation criteria, which we briefly discuss below. We also clarify a few key acronyms that are typically used to define different evaluation criteria here.
\begin{itemize}
    \item True Positive (TP): A result that is positive as both the actual value and expected value.
    \item True Negative (TN): A result that is negative as both the actual value and expected value.
    \item False Positive (FP): A false positive occurs when a projected outcome is indicated as being positive when it is actually negative.
    \item False Negative (FN): A false negative occurs when a projected outcome is indicated as being negative when it is actually positive.
\end{itemize}

\subsubsection{Precision}\label{subsubsec3.6.0}
The ratio of accurate positive predictions and all positive predictions is known as precision. Precision is the proportion of true positives to all predicted positives, which is defined as:
\begin{equation}
    \label{precision}
    \text{Precision}=\frac{TP}{TP+FP}.
\end{equation}

\subsubsection{Recall}\label{subsubsec3.6.1}
Recall, also known as Sensitivity, estimates the ratio of the number of predicted positive samples and the actual number of positive samples, which is defined as: 
\begin{equation}
    \label{recall}
    \text{Recall}=\frac{TP}{TP+FN}.
\end{equation}

\subsubsection{Specificity}\label{subsubsec3.6.2}
Specificity is the complement of Sensitivity, which estimates the ratio of the number of predicted negative samples and the actual number of negative samples. It is defined as:
\begin{equation}
    \label{specificity}
    \text{Specificity}=\frac{TN}{TN+FP}.
\end{equation}

\subsubsection{Accuracy}\label{subsubsec3.6.3}
The proportion of accurately predicted samples among all predictions is known as accuracy, which is defined as:
\begin{equation}
    \label{accuracy}
    \text{Accuracy}=\frac{TP+TN}{TP+TN+FP+FN}.
\end{equation}

\subsubsection{F1–score}\label{subsubsec3.6.4}
The weighted average of precision and recall is the F1 score. This metric is generally more beneficial than accuracy, especially if there is an uneven class distribution. F1-score is defined as:
\begin{equation}
    \label{f1_score}
    \text{F1–score}=2\times \frac{\text{Precision} \times \text{Recall}}{\text{Precision}+\text{Recall}}.
\end{equation}

\subsubsection{Intersection over Union (IoU)}\label{subsubsec3.6.5}
IoU is typically used in segmentation accuracy estimation, which is the ratio of overlap between the bounding box around a predicted object and the bounding box around the ground truth object mask. It can be defined as:
\begin{equation}
    \label{iou}
    \text{IoU}=\frac{TP}{TP+FP+FN}.
\end{equation}

\subsubsection{Sørensen–Dice Coefficient}\label{subsubsec3.6.6}
Sørensen–Dice coefficient, or simply Dice, is another common metric used in segmentation accuracy estimation, which is defined as:
\begin{equation}
    \label{dice}
    \text{Sørensen–Dice}=\frac{2\times TP}{(2\times TP)+FP+FN}.
\end{equation}

\rd{\subsubsection{Mean Squared Error (MSE)}\label{subsubsec3.6.7}
Mean Squared Error (MSE) measures the average squared difference between predicted and actual values, which is defined as:
\begin{equation}
    \label{mse}
    \text{MSE} = \frac{1}{n} \sum_{i=1}^{n} (y_i - \hat{y}_i)^2,
\end{equation}
where $n$ is the number of samples, $y_i$ represents the true value, and $\hat{y}_i$ represents the predicted value.}

%%------------------------------\\   Section 4   //------------------------------

\section{Studies using POCUS Dataset}\label{sec4}
We discussed the POCUS dataset~\citep{born2020pocovid} in Section~\ref{subsec2.1}, which can be used in making breakthroughs in the diagnosis, monitoring, and reporting of COVID-19 pneumonia in patients. This dataset contains COVID-19 (35\%), bacterial pneumonia (28\%), viral pneumonia (2\%), and healthy (35\%) classes.
\begin{table}[!ht]
\begin{center}
\begin{minipage}{350pt}
\caption{A summary of studies that used the POCUS dataset. \xmark~indicates either absent or not discussed in the article, and \checkmark~indicates present but not discussed in the article. Acronyms- ROC: Receiver Operative Characteristic, AUC: Area Under the Curve.}
\label{tab:POCUS}
\begin{adjustbox}{max width=\textwidth}

\begin{tabular}{l c c c c c c c}

\toprule
\textbf{Studies} & \textbf{AI} & \textbf{Loss} & \textbf{Results}  & \textbf{Cross-validation} 
& \textbf{Augmentation/} & \textbf{Prediction} & \textbf{Code}  \\
 &  \textbf{models} &  &  &  & \textbf{Pre-processing} & \textbf{Classes} &\\
\midrule

\citet{al2021covid} & \makecell{ResNet-18, RestNet-50, \\NASNetMobile, GoogleNet, SVM}  &  \makecell{Categorical cross-entropy} &   Accuracy: 99\%  &  k=5  &   \xmark  &   \makecell{COVID-19, CAP, Healthy}  &   \xmark \\[0.75cm]

\citet{alzogbi2021}  & DenseNet   & L1  & \makecell{Mean Euclidean \\error 14.8$\pm$7.0 mm} & \xmark & \xmark & - & \xmark\\[0.75cm]

\citet{almeida2020}    & \makecell{MobileNet}   & Categorical cross-entropy  & Accuracy: 95-100\%  & \xmark & \xmark & \makecell{Abnornal, B-lines, Mild B-lines, \\Severe B-lines, Consolidations, \\Pleural thickening} & \xmark\\[0.75cm]

\citet{awasthi2021} & \makecell{Modified MobileNet,\\CNN, and other\\lightweight models} & Focal loss & Accuracy 83.2\% & k=5 & \xmark & \makecell{COVID-19, CAP, Healthy} & \xmark\\[0.75cm]

\citet{barros2021}    & \makecell{POCOVID-Net, DenseNet, ResNet, \\NASNet, Xception-LSTM}   & \makecell{Categorical \\cross-entropy}   & \makecell{Accuracy: 93\%, \\Sensitivity: 97\%} & k=5 & \xmark & \makecell{COVID-19, Bacterial\\ Pneumonia, Healthy} & \href{https://github.com/bmandelbrot/pulmonary-covid19}{Available}\footnote{https://github.com/bmandelbrot/pulmonary-covid19}\\[0.75cm]

\citet{born2020pocovid}    & POCOVID-Net   & \makecell{Categorical \\cross-entropy}  & \makecell{AUC: 0.94,\\Accuracy: 0.89,\\Sensitivity: 0.96,\\Specificity: 0.79,\\F1-score: 0.92} & k=5 & \makecell{Rotations of up to 10\textdegree;\\Horizontal and vertical flipping;\\Shifting up to 10\% of\\the image height or width} & \makecell{COVID-19, CAP, Healthy} & \xmark\\[0.75cm]

\citet{DBLP:journals/corr/abs-2009-06116} & VGG-16 & \makecell{Categorical \\cross-entropy} & \makecell{Sensitivity: 0.98$\pm$0.04,\\ specificity: 0.91$\pm$0.08} & k=5 & \makecell{Horizontal and vertical flips, \\rotations up to 10\textdegree \\and translations of up to 10\%} & \makecell{COVID-19, CAP, Healthy} & \xmark \\[0.75cm]

\citet{born2021accelerating}  & \makecell{Frame based: VGG-16\\Video-based: Models Genesis} & \makecell{Categorical \\cross-entropy} & \makecell{Sensitivity: 0.90$\pm$0.08, \\specificity: 0.96$\pm$0.04} & k=5 & \makecell{Resizing to 224$\times$224 pixels;\\Horizontal and vertical flips;\\Rotation up to 10\textdegree;\\Translations of up to 10\%} & \makecell{COVID-19, CAP, Healthy} & \href{https://github.com/BorgwardtLab/covid19_ultrasound}{Available}\footnote{https://github.com/BorgwardtLab/covid19\_ultrasound}\\[0.75cm]

\citet{diaz2021}    & \makecell{InceptionV3, ResNet-50,\\VGG-19, Xception}  & Cross-entropy & \makecell{Accuracy: 89.1\%,\\ROC-AUC: 97.1\%} & k=5 & \makecell{Rotations (10\textdegree), horizontal\\ and vertical flips, shifts (10\%),\\ and zoom
(zoom range of 20\%)} & COVID-19, non-COVID  & \xmark\\[0.75cm]

\citet{gare2021dense} & \makecell{U-Net (reverse-transfer\\learning; segmentation\\to classification)}   & Cross-entropy  & \makecell{mIoU: 0.957$\pm$0.002,\\Accuracy: 0.849,\\Precision: 0.885,\\Recall: 0.925,\\F1-score: 0.897} & k=3 & \makecell{Left-to-right flipping;\\Scaling grey image pixels;} & \makecell{COVID-19, CAP, Healthy} & \xmark\\[0.75cm]

\citet{hou2020}  & \makecell{Saab transform based\\ successive subspace\\CNN model}   & \makecell{Categorical \\cross-entropy}  & Accuracy: 0.96  & \xmark & \makecell{Saab transformation} & \makecell{A-line, B-line,\\Consolidation} & \xmark\\[0.75cm]

\citet{karar2021lightweight}    & \makecell{MobileNets, ShuffleNets,\\ MENet, MnasNet} & \makecell{Categorical \\cross-entropy} & Accuracy: 99\% & \checkmark & \makecell{Grayscale conversion} & \makecell{COVID-19, Bacterial\\ Pneumonia, Healthy} & \xmark\\[0.75cm]

\citet{karar2021adversarial}    &  \makecell{A semi-supervised GAN, and \\a modified AC-GAN with \\auxiliary classifier} & \makecell{Min-Max loss: special\\form of cross-entropy} &  \makecell{Accuracy: 91.22\%} &  \checkmark  & \makecell{Grayscale conversion} & \makecell{COVID-19, CAP, Healthy} & \xmark\\[0.75cm]

\citet{karnes2021adaptive}    &   \makecell{Few-shot learning (FSL) visual \\classification algorithm} & Mahalanobis distances  &  ROC-AUC $>$ 85\%  & k=10  & \xmark  & \makecell{COVID-19, CAP, Healthy} & \makecell{Available upon\\request} \\[0.75cm]

\citet{muhammad2021covid}    & CNN   & \makecell{Categorical \\cross-entropy}  & \makecell{Accuracy 91.8\%,\\Precision 92.5\%,\\Recall 93.2\%} & k=5 & \makecell{Reflection around x- and y-axes;\\Rotation by [-20\textdegree, +20\textdegree];\\Scaling by a factor [0.8, 1.2]} & \makecell{COVID-19, CAP, Healthy} & \xmark\\[0.75cm]

\citet{sadik2021}    & \makecell{DenseNet-201, ResNet-152V2,\\Xception, VGG-19,\\NasNetMobile}   & \makecell{Categorical \\cross-entropy} & \makecell{Accuracy: 0.906 \\(with SpecMEn),\\F1-score: 0.90} & \checkmark & \makecell{Contrast-Limited Adaptive\\Histogram Equalization} & \makecell{COVID-19, CAP, Healthy} & \xmark\\[0.75cm]

\citet{perera2021pocformer}     &  \makecell{Transformer}    &   \makecell{Categorical \\cross-entropy}   &   Accuracy: 93.9\%     &   \checkmark     &   \xmark   & \makecell{COVID-19, CAP, Healthy} &   \xmark \\[0.75cm]

\citet{esmaeili2023covid} &  \rd{\makecell{ULBPFP-Net}} &   \makecell{Categorical \\cross-entropy}   &   \makecell{Accuracy (A): 99.76\%, \\Specificity: 99.89\%, \\Sensitivity: 99.87\%, \\F1-score: 99.75\%}  &   \checkmark     &  \makecell{Resizing to 224$\times$224 pixels;\\Isotropic transformation;\\Stratified random selection}   & \makecell{COVID-19, Viral pneumonia, Healthy} &   \xmark \\[0.75cm]

\citet{madhu2024xcovnet} &  \rd{\makecell{XCovNet}} &   \makecell{Categorical \\cross-entropy}   &   \makecell{Accuracy: 98.33\% for\\ COVID-19 detection}  &   \checkmark     &   \makecell{Use of Uniform Local Binary Pattern \\on 5 intersecting Planes (ULBPFP) to \\extract features from ultrasound data}   & \makecell{COVID-19, Lung disease, Healthy} &   \xmark \\[0.75cm]

\citet{nehary2023lung} &  \rd{\makecell{Fusion of VGG-16/ViT-generated\\ and HoG Features}} &   \makecell{Cross-entropy/Categorical \\cross-entropy}   &   \makecell{(For ViT and HoG)\\Accuracy: 91\%} &   k=10     &   \makecell{Resizing to 128$\times$128 pixels;\\ PCA for dimensionality reduction\\ of HOG features}   & \makecell{COVID-19, Bacterial Pneumonia} &   \xmark \\[0.75cm]

\citet{rahhal2022contrasting} &  \rd{\makecell{EfficinetNetB2, gMLP, ViT}} &   \makecell{Categorical cross-entropy}   &   \makecell{COVID-19 Detection\\Recall: 99.88\%,\\Precision: 95.84\%,\\F1-score: 97.41\%} &   k=5     &   \makecell{Cropping to 224$\times$224 pixels}   & \makecell{COVID-19, Bacterial\\Pneumonia, Healthy} &   \xmark \\[0.75cm]

\botrule
\end{tabular}
\end{adjustbox}
%\footnotetext{Source: This is an example of table footnote. This is an example of table footnote.}
%\footnotetext[1]{Example for a first table footnote. This is an example of table footnote.}
%\footnotetext[2]{Example for a second table footnote. This is an example of table footnote.}
\end{minipage}
\end{center}
\end{table}

\subsection{Studies}\label{subsec4.1}
In Table~\ref{tab:POCUS}, we summarize studies that used the POCUS dataset to develop and evaluate AI methods. \citet{al2021covid} utilized a set of pre-trained CNN models, namely ResNet-18, ResNet-50, GoogleNet, and NASNet-Mobile, to extract the features from the images. These features are then fed to an SVM classifier to classify the images into COVID-19, CAP, and healthy classes. A regression task was performed by~\citet{alzogbi2021}, who employed DenseNet to approximate the position of the ultrasound probe in the desired scanning areas of the torso. \citet{almeida2020} investigated a lightweight neural network, MobileNets, in the context of computer-aided diagnostics and classified ultrasound videos among abnormal, B-lines, mild B-lines, severe B-lines, consolidations, and pleural thickening classes. \citet{awasthi2021} also focused on lightweight networks that can operate on mobile or embedded devices to enable rapid bedside detection without additional infrastructure. Their method classified ultrasound images into COVID-19, CAP, and healthy classes. \citet{barros2021} proposed a CNN-LSTM hybrid model for the classification of lung ultrasound videos among COVID-19, bacterial pneumonia, and healthy classes. The extraction of the spatial feature was performed by CNNs, while the time dependency was established using the LSTM module. Born et al. published three consecutive articles~\citep{born2020pocovid,DBLP:journals/corr/abs-2009-06116,born2021accelerating} using POCOVID-Net, VGG-16, and Model-genesis, respectively, to classify lung ultrasound images into COVID-19, CAP, and healthy classes. Several pre-trained neural networks such as VGG-19, InceptionV3, Xception, and RestNet-50 have been fine-tuned on the lung ultrasound image by~\citet{diaz2021} to detect COVID-19 in the lung ultrasound test data. \citet{gare2021dense} used reverse transfer learning in a U-Net, where weights were pre-trained for segmentation and then transferred for the COVID-19, CAP, and Healthy ultrasound image classification task. In order to address the need for a less complex, power efficient, and less expensive solution to screen lung ultrasound images and monitor lung status, \citet{hou2020} introduced a Saab transform-based subspace learning model to find the A-line, B-line, and consolidation in lung ultrasound data. \citet{karar2021lightweight} introduced a lightweight deep model, COVID-LWNet, to make an efficient CNN-based system for classifying lung ultrasound images into COVID-19, bacterial pneumonia, and healthy classes. In addition, \citet{karar2021adversarial} proposed a GAN to perform the same task on ultrasound images. Few-shot learning is a machine learning framework where a machine learning model is trained with supervision using a few training samples. \citet{karnes2021adaptive} used the few-shot learning on the POCUS dataset and classified test images into COVID-19, CAP, and healthy classes. A few other approaches also used state-of-the-art CNNs~\citep{muhammad2021covid,sadik2021} or transformers~\citep{perera2021pocformer} to classify lung ultrasound images into COVID-19, CAP, and healthy classes. \rd{\citet{esmaeili2023covid} introduces ULBPFP-Net, a model combining handcrafted Uniform Local Binary Pattern on Five intersecting Planes (ULBPFP) and VGG16-generated features from lung ultrasound data. The model is designed to diagnose COVID-19, overcoming the challenges of noisy and low-quality ultrasound images. The use of five proposed planes helps capture distinctive disease features. \citet{madhu2024xcovnet} presents XCovNet, an optimized Xception-based CNN designed for diagnosing COVID-19 using POCUS images. The model employs depth-wise separable convolutions, which reduce computational complexity while maintaining high performance. The study emphasizes high accuracy and efficient feature extraction. \citet{nehary2023lung} proposes a fusion method that combines VGG-16 and ViT-generated features with a Histogram of Oriented Gradients (HoG) to classify COVID-19 and bacterial pneumonia. Dimensionality reduction is achieved using principal component analysis (PCA) on HoG features, improving classification efficiency. Finally, \citet{rahhal2022contrasting} compared the performance of EfficientNetB2, gMLP, and ViT models for COVID-19 detection. Their approach leverages these advanced architectures to achieve impressive recall and precision metrics, effectively distinguishing between COVID-19, bacterial pneumonia, and healthy cases.}

\subsection{Evaluation}\label{subsec4.2}
Studies using the POCUS dataset reported impressive results across various metrics and methodologies. For instance, \citet{al2021covid} achieved accuracy, precision, and F1-score of above 99\%. \citet{awasthi2021} developed a power and memory-efficient network that attained an impressive highest accuracy of 83.2\%. Among pre-trained models, \citet{diaz2021} found that the InceptionV3 model had the highest accuracy of 89.1\% and ROC-AUC of 97.1\%. In semantic segmentation, \citet{gare2021dense} reported high scores for various metrics, including mIoU (0.957), accuracy (0.849), precision (0.885), recall (0.925), and F1-score (0.897). Saab transform-based successive subspace learning model was reported to have an accuracy of 0.96 by \citet{hou2020}. Additionally, modified AC-GAN (accuracy: 99.45\%) outperformed semi-supervised GAN (accuracy: 99\%) in a study by~\citet{karar2021adversarial}, while MnasNet achieved the best accuracy of 99\% among six pre-trained networks. \citet{muhammad2021covid} obtained high scores for accuracy, precision, and recall (91.8\%, 92.5\%, and 93.2\%, respectively) with a fusion-based less complex CNN architecture. Real-time mass COVID-19 testing by~\citet{perera2021pocformer} resulted in over 90\% accuracy, while spectral mask enhancement (SpecMEn) improved the accuracy score of DenseNet-201 from 89.5\% to 90.4\% in a study by~\citep{sadik2021}. \rd{The ULBPFP-Net approach~\citep{esmaeili2023covid} achieved an exceptional COVID-19 detection accuracy of 99.76\%, demonstrating robustness to noise in LUS images. Specificity and sensitivity were equally impressive, reaching 99.89\% and 99.87\%, respectively, with a high F1-score of 99.75\%. Similarly, XCovNet~\citep{madhu2024xcovnet} outperformed state-of-the-art deep learning models in COVID-19 classification, delivering a strong accuracy of 98.33\%. \citet{nehary2023lung}'s fusion of ViT-generated features with HoG attained a solid accuracy of 91\%, effectively classifying COVID-19 and bacterial pneumonia. Lastly, \citet{rahhal2022contrasting} achieved a remarkable recall of 99.88\%, with a precision of 95.84\%, and an F1-score of 97.41\%, underlining the robustness of their model in distinguishing between COVID-19, bacterial pneumonia, and healthy subjects.}

\section{Studies using ICLUS-DB Dataset}\label{sec5}
We discussed the ICLUS-DB dataset in Section~\ref{subsec2.1}, which can also be used in making breakthroughs in the diagnosis, monitoring, and reporting of COVID-19 pneumonia in patients. This resource may enable AI in the identification of the disease's progress, rate, and response to treatment, facilitating more effective and personalized patient care. This dataset contains lung ultrasound data with different COVID-19 severity scores, defined as score 0: Continuous A-line (34\% of the total data), score 1: alteration in A-line (24\% of the total data), score 2: small consolidation (32\% of the total data), and score 3: large consolidation (10\% of the total data). The following table (Table \ref{tab:ICLUS-DB}) summarizes the literature on the detection of COVID-19 through the use of the ICLUS-DB dataset. 

\begin{table}[!ht]
\begin{center}
\begin{minipage}{350pt}
\caption{A summary of studies that used the ICLUS-DB dataset. \xmark~indicates either absent or not discussed in the article, and \checkmark~indicates present but not discussed in the article.}
\label{tab:ICLUS-DB}
\begin{adjustbox}{max width=\textwidth}

\begin{tabular}{@{}l c c c c c c c@{}}
\toprule
\textbf{Studies} & \textbf{AI} & \textbf{Loss} & \textbf{Results}  & \textbf{Cross-validation}
& \textbf{Augmentation/} & \textbf{Prediction} & \textbf{Code}  \\
&  \textbf{models} &  &  &  & \textbf{pre-processing} & \textbf{Classes} & \\
\midrule
    
\citet{carrer2020automatic} & HMM, VA, SVM & \xmark & \makecell{Accuracy: \\88\% (convex probe)\\ 94\% (linear probe)}   & k=10 & \xmark   & \makecell{Severity Score \\(0, 1, 2, 3)} & \xmark\\[1cm]
    
\citet{che2021multi} & \makecell{Multi-scale residual CNN}  &  Cross-entropy & \makecell{Accuracy: 95.11\%, \\F1-score: 96.70\%}  & k=5  & \makecell{Generation of local \\phase filtered and \\radial symmetry \\transformed images}  & \makecell{COVID-19, \\non-COVID}  & \xmark \\[1cm]
        
\citet{dastider2021integrated} & \makecell{Autoencoder-based\\ Hybrid CNN-LSTM} &   \makecell{Categorical \\cross-entropy}  & \makecell{Accuracy:\\67.7\% (convex probe) \\79.1\% (linear probe)} & k=5   & \makecell{Rotation, horizontal \\and vertical shift, \\scaling, horizontal\\ and vertical flips} & \makecell{Severity Score \\(0, 1, 2, 3)}  & \href{https://github.com/ankangd/HybridCovidLUS}{Available}\footnote{https://github.com/ankangd/HybridCovidLUS} \\[1cm]
    
\citet{frank2021integrating} & \makecell{ResNet-18, ResNet-101, \\VGG-16, MobileNetV2, \\MobileNetV3, DeepLabV3++} &    \makecell{SORD, \\cross-entropy}  & \makecell{Accuracy: 93\%, \\F1-Score: 68.8\%} & \xmark   & \makecell{Affine transformations, \\rotation, scaling, \\horizontal flipping,\\ random jittering} & \makecell{Severity Score \\(0, 1, 2, 3)} & \xmark \\[1cm]

\citet{roy2020deep} & \makecell{Spatial Transformer Network \\(STN), U-Net, U-Net++,\\ DeepLabV3, Model Genesis}  &  \makecell{SORD,\\ cross entropy} & \makecell{Accuracy: 96\%, \\ F1-score: 61$\pm$12\%, \\Precision: 70$\pm$19\%,\\Recall: 60$\pm$7\%}  & k=5  & \checkmark  &  \makecell{Severity Score \\(0, 1, 2, 3)} & \xmark\\[1cm]
    
\citet{khan2022deep}    & \makecell{Pre-trained CNN \\from \citep{roy2020deep}} & \makecell{SORD, \\cross-entropy}  & \makecell{Agreement-based \\scoring (82.3\%)}  & \xmark  & \xmark   & \makecell{Severity Score \\(0, 1, 2, 3)}   & \xmark \\[1cm]

\citet{khan2023benchmark}  & \rd{\makecell{ResNet-18, ResNet-50, ResNet-101, \\RegNetX, DenseNet-121, DenseNet-201, \\EfficientNetB7, and InceptionV3}} & {Cross-entropy}  & \makecell{(ResNet-18) F1-score: 65.9\% (frame-level). \\Agreement-based scoring: 59.51\% (video level), \\63.29\% (exam level), and \\84.90\% (prognostic level)}  & \xmark  & \makecell{Image cropping, removal of redundant \\information, data transformation (e.g., \\elastic warping, scaling, blurring, rotation)}  & \makecell{Severity Score \\(0, 1, 2, 3)}   & \xmark \\[1cm]

\citet{torti2024gpu} & \rd{\makecell{ResNet-50+K-means}} & \xmark  & \makecell{Accuracy: 86.2\% (binary), \\84.3\% (three-way classification), \\and 74.3\% (four-way classification)}  & \xmark  & \makecell{Frames extracted and normalized \\using OpenCV; color parameters \\converted to grayscale; ResNet-50 \\used for feature extraction} & \makecell{Severity Score \\(0, 1, 2, 3)}   & \xmark \\[1cm]

\citet{custode2023multi} & \rd{\makecell{STN, U-Net+DeepLabV3+,\\ Decision Tree}} & MSE  & \makecell{Comparable or better \\performance than state-of-the-art\\ DL methods.}  & k=5  & \xmark   & \makecell{Severity Score \\(0, 1, 2, 3)}   & \href{https://gitlab.com/leocus/neurosymbolic-covid19-scoring}{Available}\footnote{https://gitlab.com/leocus/neurosymbolic-covid19-scoring} \\[1cm]

\botrule
\end{tabular}
\end{adjustbox}
\end{minipage}
\end{center}
\end{table}

\subsection{Studies}\label{subsec5.1}
In Table~\ref{tab:ICLUS-DB}, we summarize studies that used the ICLUS-DB dataset~\citep{soldati2020proposal} to develop and evaluate AI methods. \citet{carrer2020automatic} proposed an automatic and unsupervised method to locate the pleural line using the HMM and VA. Afterward, the localized pleural line is used in a supervised SVM to classify the lung ultrasound image into COVID-19 severity scores 0-3. \citet{che2021multi} extracted local phase and radial symmetry features from lung ultrasound images, which were then fed to a multi-scale residual CNN to classify the image between COVID-19 and non-COVID classes. \citet{dastider2021integrated} incorporated an LSTM module in DenseNet-201 to predict the COVID severity between 0 and 3 in lung ultrasound images. \citet{frank2021integrating} incorporated domain-based knowledge such as anatomical features and pleural and vertical artifacts in conventional CNNs (i.e., ResNet-18, ResNet-101, VGG-16, MobileNetV2, MobileNetV3, and DeepLabV3++) to detect the severity of COVID-19 in lung ultrasound images. \citet{roy2020deep} trained several benchmark CNN models such as U-Net, U-Net++, DeepLabV3, and model genesis, incorporating STNs to simultaneously predict COVID-19 severity scores as well as localize pathological artifacts in a weakly-supervised way in the lung ultrasound images. In a unique study, \citep{khan2022deep} evaluated the performance of AI deep models in COVID-19 severity scoring by varying the image resolution and gray-level intensity of lung ultrasound images. \rd{\citet{khan2023benchmark} presented a comparative study utilizing multiple deep learning architectures, including ResNet and DenseNet variants, to assess the accuracy of frame-, video-, exam-, and prognostic-level COVID-19 severity scoring from ultrasound images. This study highlights the importance of data augmentation techniques like elastic warping and scaling to improve model performance. \citet{torti2024gpu} applied ResNet-50 combined with K-means clustering for COVID-19 severity classification, achieving competitive accuracy across binary, three-way, and four-way classification tasks, with the frames pre-processed through grayscale conversion and normalization. Lastly, \citet{custode2023multi} proposed a hybrid neuro-symbolic model integrating STN, U-Net+DeepLabV3+, and decision trees for COVID-19 severity assessment, showing comparable or superior results to state-of-the-art deep learning approaches.}

\subsection{Evaluation}\label{subsec5.2}
Studies that used the ICLUS-DB, as summarized in Table~\ref{tab:ICLUS-DB}, reported impressive results across various metrics. \citet{carrer2020automatic} reported an accuracy of 88\% and 94\% for lung ultrasound images acquired with the convex and linear probes, respectively, when they used SVM in detecting pleural line alterations due to COVID-19. \citet{che2021multi} reported an accuracy of 95.11\% and an F1-score of 96.70\% in predicting the COVID-19 severity scores in lung ultrasound. Other studies mostly predicted the COVID-19 severity scores [0, 3] using the ICLUS-DB lung ultrasound dataset as summarized in Table~\ref{tab:ICLUS-DB}. For example, accuracy in severity scoring is reported to be 67.7-79.1\%, 93\%, 96\%, and 82.3\% by \citet{dastider2021integrated}, \citet{frank2021integrating}, \citet{roy2020deep}, and \citet{khan2022deep}. \rd{ \citet{khan2023benchmark} expanded the evaluation of severity scoring by benchmarking various deep learning models, achieving a frame-level F1-score of 65.9\% with ResNet-18 and agreement-based scoring from 59.51\% at the video level to 84.90\% at the prognostic level. \citet{torti2024gpu} reported competitive accuracies in binary, three-way, and four-way classification tasks, achieving 86.2\%, 84.3\%, and 74.3\%, respectively, highlighting the model's robustness across different classification scales. Lastly, \citet{custode2023multi} demonstrated comparable or superior performance to state-of-the-art models in COVID-19 severity assessment, integrating deep learning and decision trees for an interpretable and effective analysis.
}

\section{Studies using COVIDx-US Dataset}\label{sec6}
The COVIDx-US is another large public dataset (discussed in Section~\ref{subsec2.1}) that has been thoroughly reviewed, analyzed, and validated to develop and assess AI models and algorithms~\citep{ebadi2022}. Table \ref{tab:COVIDx} summarizes existing deep learning approaches that used this dataset for COVID-19 identification and characterization in lung ultrasound images.

\begin{table}[!ht]
\begin{center}
\begin{minipage}{350pt}
\caption{A summary of studies that used the COVIDx-US dataset. \xmark~indicates either absent or not discussed in the article. Acronyms- ROC: Receiver Operative Characteristic, AUC: Area Under the Curve.}
\label{tab:COVIDx}
\begin{adjustbox}{max width=\textwidth}

\begin{tabular}{@{}l c c c c c c c@{}}
\toprule
\textbf{Studies} & \textbf{AI} & \textbf{Loss} & \textbf{Results}  & \textbf{Cross-validation}
& \textbf{Augmentation/} & \textbf{Prediction} & \textbf{Code}  \\
&  \textbf{models} &  &  &  & \textbf{pre-processing} & \textbf{Classes} & \\
\midrule

\citet{adedigba2021deep} & \makecell{SqueezeNet, \\MobileNetV2}   & \makecell{Categorical \\cross-entropy}  & \makecell{Accuracy: 99.74\%, \\Precision: 99.58\%,\\ Recall: 99.39\%} & \xmark  & \makecell{Rotation,\\ Gaussian blurring,\\random zoom,\\random lighting,\\random warp}  & \makecell{COVID-19, CAP, \\Normal, Other} & \xmark \\[1cm]

\citet{azimi2022covid}   & InceptionV3, RNN   & \makecell{Cross-entropy}  & \makecell{Accuracy: 94.44\%} & \xmark & Padding & \makecell{Positive (COVID-19), \\Negative (non-COVID-19)} & \href{https://github.com/lindawangg/COVID-Net}{Available}\footnote{https://github.com/lindawangg/COVID-Net} \\[1cm]

\citet{maclean2021covid} &  \makecell{COVID-Net US}  & \makecell{Cross-entropy}  & ROC-AUC: 0.98  &   \xmark  &   \xmark  &   \makecell{Positive (COVID-19) \\ Negative (non-COVID-19)}  & \href{https://github.com/maclean-alexander/COVID-Net-US/}{Available}\footnote{https://github.com/maclean-alexander/COVID-Net-US/} \\[1cm]

\citet{maclean2021initial}    & ResNet   & \makecell{Categorical \\cross-entropy}  & Accuracy: 0.692  & \xmark & \xmark  & \makecell{Lung ultrasound \\severity score \\(0, 1, 2, 3)} & \xmark \\[1cm]

\citet{zeng2022}    & \makecell{COVID-Net US-X}   & \makecell{Cross-entropy}  & \makecell{Accuracy: 88.4\%,\\AUC: 93.6\%} & \xmark & \makecell{Random projective \\augmentation} & \makecell{Positive (COVID-19) \\ Negative (non-COVID-19)} & \xmark\\[0.5cm]

\citet{zeng2024covid}  & \rd{\makecell{COVID-Net L2C-ULTRA}} &  \xmark  & \makecell{Accuracy: 90.6\%,\\AUC: 97.1\%,\\Recall: 97.1\%,\\Precision: 90.8\%} & \xmark & \makecell{Random projection,\\piecewise affine transformation,\\linear-convex transformation} & \makecell{Positive (COVID-19) \\ Negative (non-COVID-19)} & \xmark\\[0.5cm]

\citet{song2023covid} & \rd{\makecell{COVID-Net USPro}} & \makecell{Cross-entropy}  & \makecell{Accuracy: 99.55\%,\\Recall: 99.93\%,\\Precision: 99.83\%} & \xmark & \makecell{Resizing to $224\times224$ pixels;\\ rotation of $90^\circ$, $180^\circ$, and $270^\circ$} & \makecell{Positive (COVID-19) \\ Negative (non-COVID-19)} & \href{https://github.com/ashkan-ebadi/COVID-Net-USPro}{Available}\footnote{https://github.com/ashkan-ebadi/COVID-Net-USPro} \\[0.5cm]
\botrule
\end{tabular}
\end{adjustbox}
\end{minipage}
\end{center}
\end{table}

\subsection{Studies}\label{subsec6.1}
We summarize studies that used the COVIDx-US dataset to develop and evaluate AI methods in Table~\ref{tab:COVIDx}. \citet{adedigba2021deep} used computation and memory efficient SqueezeNet and MobileNetV2 to classify lung ultrasound images in COVID-19, CAP, normal, and other classes. Using a hybrid network consisting of the InceptionV3 model to extract spatial information and recurrent neural network (RNN) for extracting temporal features, \citet{azimi2022covid} did binary classification of lung ultrasound images into COVID-19 and non-COVID classes. \citet{maclean2021covid} proposed a DNN, COVID-Net US, leveraging a generative synthesis process that finds an optimal macro-architecture design in classifying lung ultrasound images into COVID-19 and non-COVID classes. Furthermore, \citet{maclean2021initial} used ResNet to classify lung ultrasound images into one of the four lung ultrasound severity scores (i.e., 0, 1, 2, 3). \citet{zeng2022} proposed an improved COVID-Net US network, called COVID-Net US-X, that leveraged a projective transformation-based augmentation to transform linear probe data to better resemble convex probe data. This approach performed binary classification of lung ultrasound images into COVID-19 and non-COVID classes. \rd{\citet{zeng2024covid} introduced COVID-Net L2C-ULTRA, an advanced model employing multiple transformation techniques such as random projection, piecewise affine, and linear-convex transformation. The network achieves high-performance metrics in classifying lung ultrasound images into COVID-19 and non-COVID categories. \citet{song2023covid} developed COVID-Net USPro, which classifies lung ultrasound images into COVID-19 and non-COVID categories. This model incorporates image preprocessing techniques like resizing and rotation and demonstrates exceptional performance.
}

\subsection{Evaluation}\label{subsec6.2}
The COVIDx-US dataset was used to implement various models, whose performance is illustrated by various evaluation metrics in Table~\ref{tab:COVIDx}. The models implemented by \citet{adedigba2021deep} achieved high levels of accuracy (99.74\%), precision rate (99.58\%), and recall (99.39\%). Meanwhile, \citet{azimi2022covid}'s hybrid network attained an overall accuracy of 94.44\% and learned to categorize COVID-19 as a binary classification problem. \citet{maclean2021covid}'s deep model achieved an area-under-the-curve (AUC) of over 0.98 while reducing architectural and computational complexity and inference times significantly. The ResNet implemented by \citet{maclean2021initial} achieved a total accuracy of 69.2\% with varying sensitivity values for different classes. Among all the models, the MobileNet and SqueezeNet variations of CNN performed the best on this dataset, with \citet{zeng2022} achieving a gain of 5.1\% in test accuracy and 13.6\% in AUC. \rd{COVID-Net L2C-ULTRA by \citet{zeng2024covid} achieved remarkable performance with an accuracy of 90.6\%, AUC of 97.1\%, recall of 97.1\%, and precision of 90.8\%, marking a significant improvement in classification efficiency for LUS. \citet{song2023covid} also reported exceptional results with their COVID-Net USPro model, achieving an accuracy of 99.55\%, recall of 99.93\%, and precision of 99.83\% in COVID vs. non-COVID classification, demonstrating near-perfect classification performance on LUS.}

\rd{
\section{Other Publicly Accessible LUS COVID-19 Datasets}
This section discusses three additional public datasets: Boston Emergency Department Lung UltraSound (BEDLUS), COVID-19 Simulated and Labeled \textit{In Vivo} Dataset (CSLID), Fictional Lumen Dissection Dataset (FLDD), and Lung Ultrasound COVID Phantom Dataset (LUCPD). Despite public access, to our knowledge, these datasets are used by one study each to date. Table \ref{tab:other-public-datasets} summarizes the AI studies that used these datasets.
}

\begin{table}[!ht]
\begin{center}
\begin{minipage}{350pt}
\caption{\rd{A summary of studies used the BEDLUS, CSLID, FLDD, and LUCPD datasets. \xmark~indicates either absent or not discussed in the article. Acronyms- AUC: Area Under the Curve.}}
\label{tab:other-public-datasets}
\begin{adjustbox}{max width=\textwidth}

\begin{tabular}{@{}l c c c c c c c c@{}}
\toprule
\textbf{Studies} & \textbf{Dataset} & \textbf{AI} & \textbf{Loss} & \textbf{Results}  & \textbf{Cross-validation}
& \textbf{Augmentation/} & \textbf{Prediction} & \textbf{Code}  \\
& & \textbf{models} &  &  &  & \textbf{pre-processing} & \textbf{Classes} & \\
\midrule

\citet{lucassen2023deep} &  BEDLUS  & \makecell{3D U-Net, ResNet3D-18,\\ ResNet(2+1)D-18,\\ DenseNet-121,\\EfficientNetB0, ViT,\\ DeepLabV3+}   & \makecell{Cross-entropy}  & \makecell{AUC: 95.5\%; F1-score\\ for single-point B-line\\ localization: 65\%} & k=5 & \makecell{Random translation,\\rotation, scaling, \\flipping, occlusions, \\random brightness,\\ contrast change, Gaussian noise\\ and blurring, \\resizing to 384$\times$256 pixels}  & \makecell{B-lines,\\No B-lines} & \href{https://github.com/RTLucassen/B-line_detection}{Available}\footnote{https://github.com/RTLucassen/B-line\_detection} \\[1.5cm]

\citet{zhao2024detection} &  CSLID  & \makecell{U-Net}  & Dice Loss  & \makecell{Maximum Test Dice:\\ 0.741$\pm$0.185} & \makecell{k=3} & \makecell{CSLID Data itself\\ is used as\\ augmentated data} & \makecell{B-line\\pixels } & \href{https://gitlab.com/pulselab/covid19}{Available}\footnote{https://gitlab.com/pulselab/covid19} \\[1cm]

\citet{vinod2024prognosis} & FLDD &  \makecell{GAN+Random Forest}  & -  & \makecell{(COVID-19 Detection)\\Recall: 98\%,\\Precision: 97\%,\\F-score: 98\%} &   k=5, 10  &   \makecell{Resizing to 512$\times$512 \\pixels, RGB conversion}  &   \makecell{COVID-19, CAP, \\Healthy}  &  \xmark\\[1cm]

\citet{howell2024deep} & LUCPD &  \makecell{Lightweight U-Net}  & \makecell{Combo Loss =\\ (2$\times$Dice Loss + Cross-entropy)}  & \makecell{Mean Test\\ Dice: 0.74$\pm$0.02} &   \xmark &   \makecell{Random flipping and rotation,\\varying gain, and time-gain \\compensation, random cropping \\and padding, resizing \\to 256$\times$256}  &   \makecell{Pixels for background,\\rib, pleural line,\\A-line, B-line, and \\B-line confluence}  &  \href{https://github.com/ljhowell/LUS-Segmentation-RT}{Available}\footnote{https://github.com/ljhowell/LUS-Segmentation-RT}\\[1cm]

\botrule
\end{tabular}
\end{adjustbox}
\end{minipage}
\end{center}
\end{table}

\rd{
\subsection{Studies}
Using the BEDLUS dataset,~\citet{lucassen2023deep} proposed an approach leveraging several deep learning architectures such as 3D U-Net, ResNet3D-18, ResNet(2+1)D-18, DenseNet-121, EfficientNetB0, ViT, and DeepLabV3+ for B-line detection in lung ultrasound images. The models were trained using a combination of cross-entropy loss and extensive data augmentation techniques, including random translation, rotation, scaling, flipping, and various noise and brightness adjustments. Using the CSLID,~\citet{zhao2024detection} used a U-Net-based architecture designed specifically for B-line segmentation in ultrasound images. Dice loss was utilized to optimize the model, and the CSLID dataset itself served as the augmented data for training in conjunction with small-cohort real LUS data. Using the FLDD dataset,~\citet{vinod2024prognosis} employed a hybrid model combining a GAN with a Random Forest classifier to detect COVID-19 from chest imaging data, with preprocessing steps involving image resizing and RGB conversion. Additionally,~\citet{howell2024deep} used the LUCPD dataset to develop a lightweight U-Net model for lung ultrasound segmentation. The architecture was optimized using a combination of Dice loss and cross-entropy, while augmentation techniques such as random flipping, rotation, and cropping were applied to improve the robustness of the model in detecting key lung features like A-lines, B-lines, and pleural lines.

\subsection{Evaluation}
\citet{lucassen2023deep} reported an AUC of 95.5\% in B-line detection, with a moderate F1-score of 65\% for single-point B-line localization, indicating strong detection performance but challenges in precise localization. \citet{zhao2024detection} achieved a maximum test Dice score of 0.741$\pm$0.185 for B-line segmentation using the CSLID dataset, highlighting the variability in segmentation performance. Meanwhile, \citet{vinod2024prognosis} demonstrated robust results in COVID-19 detection, reporting recall and F1-scores of 98\% and precision of 97\%, showcasing the effectiveness of their approach in distinguishing COVID-19 cases from other conditions such as CAP and healthy individuals. Finally,~\citet{howell2024deep} achieved a mean test Dice score of 0.74$\pm$0.02 for segmentation of lung features such as ribs, pleural lines, and B-lines, demonstrating the proficiency of the model in identifying various structures relevant to lung ultrasound interpretation, though with some variability in performance.}

\rd{\section{Studies using Non-Accessible LUS COVID-19 Private Datasets}}
\label{sec7}
Several studies utilized privately owned datasets, which are not publicly accessible, as mentioned in section~\ref{subsec2.2}. However, some of these datasets' primary sources, such as hospitals, clinics, and online repositories, have overlapped with those of public data. Although some private dataset links could not be traced due to lack of availability in the articles, some can be accessed by sending a request for use, for example, \citet{camacho2022artificial,durrani2022automatic,rojas2021detection}.

\begin{table}[!ht]
\begin{center}
\begin{minipage}{350pt}
\caption{A summary of studies that used private datasets. \xmark~indicates either absent or not discussed in the article. Acronyms- ROC: Receiver Operative Characteristic, AUC: Area Under the Curve.}
\label{tab: private}
\begin{adjustbox}{max width=\textwidth}

\begin{tabular}{@{}l c c c c c c c@{}}
\toprule
\textbf{Studies} & \textbf{AI} & \textbf{Loss} & \textbf{Results}  & \textbf{Cross-validation}
& \textbf{Augmentation/} & \textbf{Prediction} & \textbf{Code}  \\
&  \textbf{models} &  &  &  & \textbf{pre-processing} & \textbf{Classes} & \\
\midrule

\citet{arntfield2020development}    &   Xception   &   Binary Cross Entropy  &   ROC-AUC: 0.978  &  \xmark  & \makecell{Random zooming in/out by $\leq$10\%,\\ horizontal flipping, horizontal\\ stretching/contracting by $\leq$20\%,\\ vertical stretching/contracting ($\leq$5\%),\\ and bi-directional rotation by $\leq10^\circ$}  & \makecell{Hydrostatic pulmonary \\edema (HPE), onn-COVID \\acute respiratory \\distress syndrome (ARDS),\\COVID-19 ARDS} & \href{https://github.com/bvanberl/covid-us-ml}{Available}\footnote{https://github.com/bvanberl/covid-us-ml}\\[1cm]

\citet{chen2021quantitative}    &   \makecell{2-layer NN, SVM, \\Decision Tree}    &   \xmark  &  Accuracy: 87\%  &   k=5  &   \makecell{Curve-to-linear \\ conversion}  &   \makecell{Score 0: Normal,\\Score 1: Septal syndrome,\\Score 2: Interstitial-alveolar syndrome,\\Score 3: White lung syndrome}  &  \xmark \\[1cm]

\citet{durrani2022automatic}    &   \makecell{CNN, Reguralized \\STN (Reg-STN)}    &   SORD    &   \makecell{Accuracy: 89\%, \\PR-AUC: 73\%}   &   k=10    &   \makecell{Replacing overlays, \\resizing to 806$\times$550 pixels} &   \makecell{Consolidation present, \\consolidation absent}  &   \xmark \\[1cm]

\citet{ebadi2021automated} & Kinetics-I3D & \makecell{Focal loss} & \makecell{Accuracy: 90\% \\ Precision: 95\%} & k=5 & \xmark & \makecell{A-line (normal), \\B-line, \\Consolidation and/or\\ pleural effusion} & \xmark\\[1cm]

\citet{huang2022evaluation} &   \makecell{Non-local Channel \\Attention ResNet} &  Cross-entropy  &  \makecell{Accuracy:  92.34\%, \\F1-score: 92.05\%,\\ Precision: 91.96\%\\ Recall: 90.43\%},  &  \xmark &   \makecell{Resizing to \\300$\times$300 pixels} &   \makecell{Score 0: normal, \\Score 1: septal syndrome, \\Score 2: interstitial-alveolar syndrome, \\Score 3: white lung syndrome}  &   \href{https://biohsi.ecnu.edu.cn}{Available}\footnote{https://biohsi.ecnu.edu.cn}    \\[1cm]

\citet{la2021deep}  &   ResNet-18, ResNet-50  &   Cross-entropy   &   F1-score: 98\%  &   \xmark  &   \makecell{Geometric, filtering, random centre \\cropping, and colour transformations}   &   \makecell{Severity score:\\0, 0*, 1, 1*, 2, 2*, 3}  &   \xmark    \\[1cm]

\citet{liu2020semi} &   \makecell{Multi-symptom multi-label\\(MSML) network} &   Cross-entropy   &   \makecell{Accuracy: 100\% \\(with 14.7\% data)}  &   \xmark  &   \makecell{Random rotation \\(up to 10 degrees)\\ and horizontal flips} &   \makecell{A-line, B-line,\\Pleural lesion, Pleural effusion}   &   \xmark\\[1cm]

\citet{mento2021deep}   &   STN, U-Net, DeepLabV3+   &   \xmark  &  \makecell{Agreement between \\AI scoring and\\ expert scoring 85.96\%} &   \xmark  &   \xmark  &   \makecell{Expert scores:\\0, 1, 2, 3}  &   \xmark      \\[1cm]

\citet{quentin2020extracting}   &   ResNet-18    & Cross-entropy &   Accuracy (Val): 100\%  &   \xmark    &   Resizing to 349$\times$256   &   \makecell{Ultrasound frames \\with (positive) and \\without (negative) \\clinical predictive value}  &   \xmark  \\[1cm]

\citet{nabalamba2022machine}    &   \makecell{VGG-16, VGG-19, ResNet}   &   Binary cross-entropy   &  \makecell{Accuracy: 98\%,\\ Recall: 1,\\Precision: 96\%,\\F1-score: 97.82\%,\\ ROC-AUC: 99.9\%} &   \xmark   &    \makecell{Width and height shifting, \\ random zoom within 20\%,\\ brightness variations within [0.4, 1.3],\\rotation up to 10 degrees} &   \makecell{COVID-19, Healthy}  &   \xmark  \\[1cm]

\citet{panicker2021approach}    &   LUSNet (U-Net based CNN)    &   Categorical cross-entropy   &   \makecell{Accuracy: 97\%,\\Sensitivity: 93\%,\\Specificity: 98\%}   & k=5  &   \makecell{Generation of local\\ phase and shadow back\\scatter product images}   &   \makecell{Classes: 1, 2, 3, 4, 5} &   \href{https://github.com/maheshpanickeriitpkd/ALUS}{Available}\footnote{https://github.com/maheshpanickeriitpkd/ALUS}  \\[1cm]

\citet{roshankhah2021investigating} &   U-Net &   Categorical cross-entropy   &   Accuracy: 95\%   &    \xmark    & \makecell{Randomly cropping and\\ rotating the frames}   &   \makecell{Severity score:\\0, 1, 2, 3}  &   \xmark  \\[1cm]

\citet{wang2021semi}    &   SVM  &   \xmark  &   \makecell{ROC-AUC: 0.93,\\Sensitivity: 0.93,\\Specificity: 0.85} &   \xmark  &   \xmark  &   \makecell{Non-severe, severe}  &   \xmark  \\[1cm]

\citet{xue2021modality} &   \makecell{U-Net (with  modality alignment \\contrastive learning of \\representation (MA-CLR))} &   \makecell{Dice, \\cross-entropy}  &   \makecell{Accuracy:\\75\% (4-level)\\ 87.5\% (binary)} &   \xmark  &   \makecell{Affine transformations (translation, \\rotation, scaling, shearing),\\ reflection, contrast change, Gaussian \\noise, and Gaussian filtering}    &    \makecell{Severity score:\\0, 1, 2, 3}  &   \xmark\\[1cm]

\citet{kuroda2023artificial} & \rd{\makecell{AI-POCUS (model specifics\\are not disclosed)}} &   \makecell{\xmark}  &   \makecell{(COVID-19 Detection)\\Accuracy: 94.5\% for 12-zone, 83.9\% for 8-zone\\Sensitivity: 92.3\% for 12-zone, 77.5\% for 8-zone\\Specificity: 100\% for both 12- and 8-zone} &   \xmark  &  \xmark  &    \makecell{Count of B-lines in each zone\\($\geq 3$ means abnormal)}  &   \xmark\\[1cm]

\citet{sagreiya2023automated} & \rd{\makecell{Calculated Lung \\Ultrasound (CLU)}} &   \makecell{\xmark}  &   \makecell{100\% concordance \\between CLU and radiologist\\ findings in COVID-19 severity} &   \xmark  &  \xmark  &    \makecell{Presence of A-lines, \\tiny B-lines,\\confluent B-lines,\\ pleural effusion,\\ thick B-lines, and \\B-lines with consolidations.}  &   \xmark\\[1cm]

\citet{faita2024covid} & \rd{\makecell{I3D}}  & \xmark & \makecell{Accuracy dropped from \\70.4\% (in 2020) to\\48.3\% (in 2022),\\ with increased MAE} &   \xmark  &  \makecell{Random cropping, \\horizontal flipping, \\rotation}  & \makecell{Severity score:\\0, 1, 2, 3}  & \xmark\\[1cm]

\citet{kimura2024effectiveness} & \rd{\makecell{CNN}}  & \xmark & \makecell{Accuracy: 79\% (layperson), \\Accuracy: 80\% (physician); \\Sensitivity: 84\% (laypersons), \\Sensitivity: 94\% (physicians)} &   \xmark  &  \xmark  & \makecell{Normal (no B-lines), \\Mild-moderately abnormal (1-2 B-lines), \\Severely abnormal ($\geq3$ or coalesced B-lines)}  &   \xmark\\[1cm]

\citet{li2024knowledge} & \rd{\makecell{KFLR Transformer}}  & \makecell{Categorical cross-entropy} & \makecell{Binary-level accuracy: 96.4\%, \\Four-level accuracy: 87.4\%} &   \xmark  &  \makecell{Image normalization, \\pixel size calibration, \\intensity normalization, and \\transformation into polar \\coordinates}  & \makecell{Binary-level severity,\\Four-level severity (0, 1, 2, 3)}  &   \xmark\\[1cm]
\botrule
\end{tabular}
\end{adjustbox}
\end{minipage}
\end{center}
\end{table}

\subsection{Studies}\label{subsec7.1}
\citet{arntfield2020development} highlighted the need for collaborative research involving multi-center for the discrepancy in results between the model and people, which shows the presence of hidden biomarkers within ultrasound images. In addition, they trained the Xception neural network to classify LUS images into hydrostatic pulmonary edema (HPE), non-COVID acute respiratory distress syndrome (ARDS), and COVID-19 ARDS. \citet{chen2021quantitative} employed a 2-layer neural network to extract image features, which were subsequently used in an SVM and decision tree algorithm for predicting LUS scores between 0 to 3 (i.e., score 0: normal, score 1: septal syndrome, score 2: interstitial-alveolar syndrome, and score 3: white lung syndrome). \citet{durrani2022automatic} used an autonomous deep learning-based technique to detect consolidation/collapses in LUS images. A CNN and Reg-STN-based model has been used with a SORD cross-entropy loss function. A fast and dependable interpretation of LUS images without preprocessing or frame-by-frame analysis was presented by \citet{ebadi2021automated}. They proposed a two-stream inflated 3D CNN, referred to as Kinetics-I3D, to detect A-line (normal), B-line, consolidation, and/or pleural effusion in LUS images. \citet{huang2022evaluation} proposed a non-local channel attention ResNet to facilitate extraction of the dependencies between distant pixels and stressing specific key channels. Their method classified LUS images into four scores (i.e., score 0: normal, score 1: septal syndrome, score 2: interstitial-alveolar syndrome, and score 3: white lung syndrome). \citet{la2021deep} used ResNet-18 and ResNet-50 to perform a seven-way classification of LUS images. Classes include score 0: A-lines, score 0*: A-lines not defined, score 1: an irregular or damaged pleural line along with visible vertical artifacts, score 1*: pleural line not defined, score 2: broken pleural line with either small or broad consolidated areas with wide vertical artifacts below (white lung), score 2*: broken pleural line not defined, and score 3: dense and broadly visible white lung with or without larger consolidations. \citet{liu2020semi} proposed a novel multi-symptom multi-label (MSML) network incorporating a semi-supervised two-stream active learning strategy, which detected A-line, B-line, pleural lesion, and pleural effusion in LUS images. In a different type of study, \citet{mento2021deep} estimated the agreement of the COVID-19 severity scores predicted by deep models (i.e., STN, U-Net, and DeepLabV3+) to the expert scores. \citet{quentin2020extracting} used a pre-trained ResNet-18 to automate the selection of clinically meaningful and predictive image frames from LUS videos that have high clinical predictive value. \citet{nabalamba2022machine} used three pre-trained deep learning models (i.e., VGG-16, VGG-19, and ResNet) to detect COVID-19 from LUS images. \citet{panicker2021approach} designed a U-Net for lung ultrasound image analysis, called LUSNet, which is trained to classify ultrasound images into five classes of increasing severity of regions. They have followed the rectification of the ultrasound images to make them agnostic to the type of probe employed and to restrict unwanted edge effects, particularly in the case of convex and sector probes. In a typical abnormal lung ultrasound image, B-line artifacts appear, which often evolve into white lung patterns in the more severe cases. Exploiting these anatomical changes, \citet{roshankhah2021investigating} used the U-Net-based segmentation approach to automatically stage the progression of COVID-19. While most AI approaches for COVID-19 detection and analysis adopted deep learning techniques, \citet{wang2021semi} extracted hand-engineered features such as thickness and roughness of the pleural line, and the accumulated with an acoustic coefficient of B lines, which were subsequently used in an SVM to classify lung ultrasound images into severe and non-severe cases. \citet{xue2021modality} performed a comprehensive study using the features from LUS data and clinical information in supervised attention-based multiple instance learning (DSA-MIL) modules to classify LUS images into four severity grades. \rd{\citet{kuroda2023artificial} investigated an AI-based POCUS (AI-POCUS) for detecting COVID-19 pneumonia. They used a commercially available app developed by Philips and compared the AI-POCUS findings with CT scans. The study involved 56 subjects and focused on quantifying B-lines in lung zones. The AI-POCUS system demonstrated high accuracy, even with minimal user experience, showcasing its potential as a rapid screening tool in resource-limited settings. \citet{faita2024covid} utilized 3D-based deep learning models (Inflated 3D ConvNet) on lung ultrasound videos to predict COVID-19 severity. The model was trained on a 2020 dataset and tested on a 2022 cohort to evaluate its robustness over time. The results indicated a significant drop in performance as the disease characteristics evolved, highlighting the importance of continuous model retraining. \citet{kimura2024effectiveness} explored an AI-driven audio output method to assist laypersons in recognizing pulmonary edema or COVID-19 lung infection on ultrasound images. A CNN model was trained to output audio cues, making it easier for untrained users to interpret the images. The study found that laypersons using audio cues performed comparably to physicians, suggesting its potential for self-monitoring and remote healthcare applications. \citet{li2024knowledge} introduced a novel Knowledge Fusion Latent Representation (KFLR) framework for assessing COVID-19 pneumonia severity using LUS images. Integrating clinician-guided knowledge into deep learning models improved both accuracy and interpretability. The model achieved high accuracy for both binary and multi-level severity assessments, demonstrating its effectiveness in clinical decision-making.}

\subsection{Evaluation}
\label{subsec7.2}
Various metrics have been used to evaluate the performance of methods that used private datasets. \citet{arntfield2020development} were able to distinguish between COVID-19 (AUC = 1.0), non-COVID (AUC = 0.934), and HPE (AUC = 1.0) with high AUCs, whereas Physicians' performance for COVID-19, non-COVID, and HPE detection had AUCs of 0.697, 0.704, and 0.967, respectively. \citet{camacho2022artificial} achieved high agreement between the expert and algorithm for detecting B-Lines (88.0\%), consolidations (93.4\%), and pleural effusion (99.7\%), and moderate agreement for the individual video score (72.8\%). \citet{chen2021quantitative} performed a comparison of performance by CNN, SVM, and decision tree models, where the CNN performed the best, achieving 87\% accuracy over traditional machine learning models. In the study of \citet{durrani2022automatic}, the video-based supervised learning method outperformed a fully supervised frame-based method in terms of PR-AUC, with scores of 73.34 and 60.08, respectively.
Using a classification model originally developed for recognizing human action, \citet{ebadi2021automated} achieved high accuracy (90\%) and average precision (95\%). Using a non-local channel attention ResNet, \citep{huang2022evaluation} achieved superior performance compared to conventional ResNet, VGG, and other networks, with an accuracy of 92.34\% and F1-score of 92.05\%. \citet{liu2020semi} reported 100\% accuracy for regional classification by training only 14.7\% of the data, with comparable performance in sensitivity (92.38\%) and specificity (100\%). \citet{nabalamba2022machine} also achieved an accuracy of 98\%, along with other high metrics (precision of 95.74, recall of 1.00, F1-score of 97.82\%, and ROC-AUC of 99.99\%) for the classification of patients at high risk of clinical deterioration and patients at low risk. Similarly, \citet{mento2021deep} showed a high percentage of agreement (85.96\%) for the classification of patients at high risk of clinical deterioration and patients at low risk with that by expert radiologists. \citet{quentin2020extracting} employed a transfer learning-based approach that achieved high validation accuracy (99.74\%) for data with varying brightness levels. Using deep learning approaches, higher accuracy of 97\% and 95\% are also reported in COVID-19 detection in ultrasound by \citet{panicker2021approach} and \citet{roshankhah2021investigating}, respectively. \citet{wang2021semi}, on the other hand, used an SVM classifier that achieved a good binary classification performance between severe and non-severe cases (sensitivity = 0.93, specificity = 0.85, ROC-AUC = 0.93). By combining lung ultrasound data and clinical information in a multiple instance learning framework, \citet{xue2021modality} were able to categorize patients' clinical severity into four groups with 75\% accuracy and into severe/non-severe groups with 87.5\% accuracy. \rd{\citet{kuroda2023artificial} demonstrated that AI-POCUS had a high accuracy of 94.5\% for detecting COVID-19 pneumonia when using a 12-zone scan and 83.9\% for an 8-zone scan, validated against CT findings. The study showed the potential of AI-POCUS as an effective screening tool. Still, it was limited by its small sample size, single-center design, and lack of comparison with expert-performed traditional ultrasound. \citet{faita2024covid} reported a significant decline in model accuracy, from 70.4\% in the 2020 cohort to 48.3\% in the 2022 cohort, when using 3D-based deep learning models on lung ultrasound videos. The study highlighted the need for continuous retraining due to evolving disease characteristics, with increased Mean Absolute Error (MAE) and Root Mean Square Error (RMSE) over time. \citet{kimura2024effectiveness} found that AI-generated audio outputs allowed laypersons to identify B-lines in lung ultrasound images with an accuracy of 79\%, comparable to physician accuracy of 80\%. The sensitivity of laypersons was 84\%, slightly lower than that of physicians, at 94\%. While promising, the study's small sample size and lack of video interpretation in the layperson group limited its generalizability. \citet{li2024knowledge} achieved high accuracy with their KFLR model, reporting 96.4\% accuracy for binary-level severity assessment and 87.4\% for four-level severity assessment using lung ultrasound images. The study demonstrated the method’s effectiveness in severity evaluation but was limited by its single-center dataset and reliance on static images, warranting further validation on larger datasets.
}

\rd{
\section{Challenges, Limitations, and Gaps in Research}
\label{sec8}
Despite significant advancements in the application of AI for COVID-19 detection using LUS data, several challenges and gaps continue to hinder progress and clinical adoption in this field.

\textbf{Limited Availability and Diversity of Datasets.} A critical challenge is the limited availability of large, diverse, and high-quality ultrasound datasets. Most publicly available LUS datasets are small, region-specific, and lack sufficient representation across different patient demographics, comorbidities, and disease stages. This narrow scope restricts the generalizability of AI models, making it difficult for them to perform consistently across diverse populations and clinical settings. Additionally, the lack of longitudinal datasets further limits the ability to evaluate the long-term reliability and clinical utility of AI models in detecting COVID-19.

\textbf{Lack of Standardized Methodological Reporting.} Many studies fail to provide comprehensive documentation of their methodologies, particularly regarding image preprocessing, augmentation, and model implementation. Inconsistent or incomplete reporting makes it challenging to replicate experiments and benchmark new models. The absence of standardized protocols for data handling, model training, and evaluation impedes reproducibility and limits the transparency required for scientific validation. To advance the field, the research community must prioritize the development of clear, standardized guidelines for methodological reporting.

\textbf{Limited Public Availability of AI Models and Codebases.} Another significant limitation is the scarcity of publicly available AI model codebases. While datasets are crucial, open access to code is equally important for fostering collaboration and accelerating advancements. Unfortunately, few studies share their models and codebases openly, e.g.,~\citep{barros2021,born2020pocovid,dastider2021integrated,roy2020deep,azimi2022covid}, which hinders researchers from validating, improving, or building upon existing models. To promote transparency and reproducibility, the sharing of code and detailed implementation guidelines should become a standard practice in this field.

\textbf{Lack of Explainability and Interpretability in AI Models.} While many AI models achieve impressive accuracy in detecting COVID-19 from ultrasound data, a critical gap exists in their explainability and interpretability. In medical applications, AI models must be transparent and interpretable to gain the trust of healthcare professionals and facilitate clinical adoption. Currently, few studies address the need for explainable AI (XAI) frameworks, leaving clinicians in the dark regarding how AI models make decisions. Without interpretability, it becomes difficult to integrate these models into real-world clinical workflows, where trust and understanding of AI outputs are essential.

\textbf{Insufficient Focus on Clinical Integration and Longitudinal Validation.} Although various AI models have shown high diagnostic performance, there is a lack of longitudinal studies assessing their clinical utility over time. Most models focus solely on diagnostic accuracy without considering their practical integration into existing healthcare systems. Questions remain about how these models perform in dynamic clinical environments, how they can be integrated into decision-making processes, and their ability to adapt to new challenges, such as emerging COVID-19 variants. Addressing these issues requires large-scale clinical trials and the collaboration of healthcare professionals in evaluating AI-driven ultrasound diagnostics.
}

\rd{
\section{Discussion and Future Works}
\label{sec9}
We began this survey with 874 initial search-yielded articles on the topic of COVID-19 detection using AI on LUS from Google Scholar. After several filtering phases, as discussed in section~\ref{sec1}, we reviewed a total of 60 LUS studies that focused on COVID-19 detection or analysis using AI. Some of the key observations that can be noted from this review are as follows:

\textbf{COVID-19 Severity Assessment.} LUS can be helpful in COVID-19 severity assessment in patients as supported by the studies in the survey~\citep{carrer2020automatic,che2021multi,dastider2021integrated,frank2021integrating,roy2020deep}. COVID-19 primarily affects the respiratory system, causing pneumonia and acute respiratory distress syndrome (ARDS). LUS can detect these lung abnormalities earlier than chest X-rays and provide detailed information on the extent and severity of lung involvement~\citep{martinez2021higher}. It can also help differentiate COVID-19 pneumonia from bacterial or viral pneumonia (i.e., CAP). Overall, ultrasound is a safe and non-invasive imaging modality that can provide valuable information for the assessment and management of COVID-19 in patients, especially pregnant women and children. It can help detect early lung involvement, monitor disease progression, and guide clinical decision-making.

\textbf{Data Partition for Benchmarking.} Although numerous publicly available datasets are available, studies have reported varying degrees of quantitative accuracy in detecting, segmenting, and assessing the severity of COVID-19 independently. Without replicating the results of a particular study that used a publicly available ultrasound dataset, it is impossible to make a fair comparison of methodological performance. However, this issue of complexity can be resolved by partitioning a specific portion of a publicly available dataset for quantitative validation across studies. This benchmark dataset can then be used for model validation and quantitative accuracy comparison.

\textbf{Addressing the Research Questions.} Our review paper comprehensively addressed the key research questions (RQs) related to AI-based COVID-19 detection using LUS data. RQ1 focused on identifying the most commonly used public and private LUS datasets for COVID-19 detection. We systematically reviewed and cataloged 7 unique public datasets and 21 private datasets, providing a thorough overview of dataset accessibility and availability. RQ2 examined the variation in AI-based methods applied to COVID-19 detection across different datasets, highlighting key performance metrics such as accuracy, sensitivity, and specificity. Our analysis revealed notable variations in performance metrics across datasets, which underscores the importance of dataset-specific optimization and benchmarking. RQ3 investigated common ultrasound image preprocessing and augmentation techniques used to enhance model performance. We identified and summarized prevalent methods, such as normalization, data augmentation, and denoising, and discussed their impact on model efficacy. RQ4 addressed the existing limitations and challenges in the use of ultrasound for COVID-19 detection. We highlighted issues such as the limited availability of diverse datasets, the lack of standardized methodologies, and the need for explainable AI frameworks. Our review provides actionable insights and recommendations for overcoming these challenges, emphasizing the importance of robust, transparent, and reproducible research practices. Through this comprehensive analysis, our paper fills a critical gap in the literature by offering a detailed and organized perspective on current practices, challenges, and future directions in the field of AI-based COVID-19 ultrasound diagnostics.

\textbf{Potential Future Works.} Based on the observation in this review, we foresee several research directions that can be pursued in the future:
\begin{itemize}
    \item \textit{Developing a standardized protocol for ultrasound-based severity assessment of COVID-19}: The studies in the survey highlight the potential of LUS in assessing the severity of COVID-19. However, there is a need to develop a standardized protocol for LUS-based severity assessment to ensure consistency across studies and to facilitate comparisons between different AI models. This protocol should include standardized imaging techniques, imaging parameters, and diagnostic criteria.

    \item \textit{Integration of LUS with other imaging modalities}: While LUS is a useful tool for COVID-19 assessment, it has limitations, such as limited penetration depth and difficulty in imaging certain structures. Future work can focus on combining LUS with other imaging modalities, such as CT or X-ray (if available), to provide a more comprehensive assessment of COVID-19.

    \item \textit{Integrating AI models for early detection and monitoring of COVID-19}: LUS can detect early lung involvement and monitor disease progression in COVID-19 patients. Future work can focus not only on developing but also on integrating AI models in clinical settings that can accurately detect COVID-19 at an early stage and monitor disease progression over time, enabling timely intervention and better patient outcomes.

    \item \textit{Comparison of AI models using benchmark datasets}: As highlighted in the discussion, there is a need for benchmark datasets for quantitative accuracy comparison of different AI models. Future work can focus on developing benchmark datasets and using them to compare the performance of different AI models for COVID-19 detection and analysis.

    \item \textit{Integration of AI models into clinical practice}: The potential of AI models for COVID-19 detection and analysis is vast, but their integration into clinical practice is still limited. Future work can focus on developing user-friendly and interpretable AI models that can be easily integrated into clinical workflows, improving the accuracy and speed of COVID-19 diagnosis and treatment.

    \item \textit{Exploration of novel pre-processing and augmentation techniques}: The quality of input data is crucial for the accuracy of AI models. Future work can focus on exploring novel pre-processing and augmentation techniques for ultrasound images to improve the quality of input data and the performance of AI models. These techniques can include advanced filtering, contrast enhancement, or more sophisticated augmentation methods.

    \item \textit{Integration of clinical and imaging data}: AI models for COVID-19 detection and analysis can benefit from the integration of clinical and imaging data. Future work can focus on developing AI models that can integrate clinical and imaging data to provide a more comprehensive assessment of COVID-19 and its impact on patients.

    \item \textit{Development of explainable AI (XAI) for ultrasound-based diagnosis}: Given the importance of transparency and trust in medical AI applications, future research can focus on incorporating explainable AI techniques more into LUS-based COVID-19 detection models. This could help healthcare professionals understand the underlying decision-making process of the models and improve their adoption in clinical settings.

    \item \textit{Federated learning for privacy-preserving model training}: Privacy concerns around patient data sharing remain a critical issue. Future work could explore the use of federated learning, where models are trained across decentralized healthcare institutions without sharing sensitive patient data. This could facilitate large-scale AI model training on diverse datasets without compromising privacy.

    \item \textit{Personalized AI models for COVID-19 detection}: Developing personalized AI models tailored to individual patient characteristics, such as age, gender, and pre-existing conditions, can improve diagnostic accuracy. Future work can focus on building adaptive AI models that account for individual variability in disease progression.

    \item \textit{Application of multi-task learning}: Multi-task learning could allow AI models to perform multiple related tasks simultaneously, such as COVID-19 detection, severity assessment, and segmentation of lung lesions. Research in this area could lead to more efficient and versatile models for COVID-19 diagnosis using ultrasound.

    \item \textit{AI for ultrasound image quality improvement}: Many existing studies rely on suboptimal or noisy ultrasound images. Future research could focus on leveraging AI techniques to enhance the quality and resolution of ultrasound images before they are used for diagnostic purposes, potentially improving the performance of downstream AI models.

    \item \textit{AI-powered telemedicine for remote ultrasound diagnostics}: AI models can be integrated into telemedicine platforms to enable remote diagnosis of COVID-19 using portable ultrasound devices. This future direction could help extend diagnostic capabilities to rural or underserved areas where access to advanced imaging tools is limited.

\end{itemize}}

\section{Conclusions}
\label{con}

In this comprehensive review, we provide a detailed survey of LUS-based AI COVID-19 studies that have utilized both publicly available and private LUS datasets. The main contributions of this review include an exhaustive survey of articles using publicly available LUS datasets for COVID-19, a listing and review of these datasets, and the organization of LUS-based AI studies by dataset. Additionally, this review analyzes and tabulates studies across several dimensions, such as data preprocessing, AI models, cross-validation, and evaluation criteria, and summarizes all reviewed works in a tabular format to facilitate easier comparison among studies. The search strategy employed was thorough, with a total of 60 articles reviewed, with 41 using public datasets and the remainder using private data. We selected articles based on criteria including full-text availability, use of AI techniques for COVID-19 detection or analysis from LUS data, hypothesis support through qualitative and quantitative results, and adherence to a minimum standard of quality.

\rd{However, it is important to note some limitations of our review. Despite our rigorous search and selection process, some relevant studies may have been missed due to the exclusion of non-English publications and articles not indexed in the major databases we focused on. Additionally, the variability in AI methodologies and dataset characteristics could mean that some findings may not be directly comparable. These limitations suggest that future research should aim for a more inclusive and standardized approach to enhance the generalizability of AI models for COVID-19 detection using lung ultrasound data. This review provides valuable insights into the current state of LUS-based AI COVID-19 studies and serves as a crucial resource for researchers in this field. The findings can aid in developing more accurate and efficient AI models for COVID-19 detection and diagnosis, ultimately improving patient care and outcomes.}

\section*{Declarations}
\subsection*{Ethical Approval}
This study is a comprehensive review of existing literature and did not involve any new studies of human or animal subjects performed by any of the authors. Therefore, it did not require any ethical approval or informed consent.

\subsection*{Funding}
No funding was received to conduct this study.

\subsection*{Conflict of Interests/Competing Interests}
The authors have no financial or proprietary interests in any material discussed in this article.

\subsection*{Availability of Data and Materials}
Data sharing does not apply to this article as no datasets were generated or analyzed during the current study. This study is a comprehensive review of existing literature. All human data referenced in this review are from previously published studies, and links to these studies and their datasets are provided in the manuscript. No new human data were directly used or accessed by the authors of this review.

\subsection*{Authors' contributions}
\textbf{M.A.J., M.S.I.W.}: Conceptualization, Methodology, Investigation, Writing - original draft, Visualization, Data Curating, Software, Writing - review \& editing.\\
\textbf{M.A.H.}: Conceptualization, Methodology, Investigation, Writing - original draft, Supervision, Writing - review \& editing, Project Administration, Validation.\\
\textbf{A.M., A.A.S., M.J.A.N., M.M.H.S., M.I.U., M.I.S., N.A., S.R.S., T.R.}: Conceptualization, Methodology, Formal analysis, original draft, Data Curating.\\
\textbf{M.K., M.R.D., M.H., R.S., R.C., S.B.E., T.I.}: Writing - original draft, Formal analysis, Data Curating.\\
\textbf{Equal Contribution}: M.A.J., M.S.I.W., and M.A.H. contributed equally as co-first authors.\\
\textbf{Shared Equal Contribution}: A.M., A.A.S., M.J.A.N., M.M.H.S., M.I.U., M.I.S., N.A., S.R.S., T.R.

\subsection*{Acknowledgements}
We would like to extend our deepest gratitude to the Associate Editor and anonymous reviewers for their valuable time, insightful comments, and constructive feedback. Their thoughtful suggestions have significantly contributed to improving the quality and clarity of this manuscript. We appreciate their dedication and effort in helping us refine this work.

\bibliography{ultrasound_references}

\end{document}